%% file: main.tex
\documentclass[11pt]{article}

\usepackage[final]{automl}

\input{Utils/packages}

\input{Utils/commands}

\input{Utils/macros.tex}

\usepackage{natbib}
\title{Learning Activation Functions for Sparse Neural Networks}

\author[1]{\nameemail{Mohammad Loni$^*$}{mohammad.loni@mdu.se}}
\author[2]{\nameemail{Aditya Mohan$^*$}{a.mohan@ai.uni-hannover.de}}
\author[3]{\nameemail{Mehdi Asadi}{mehdi.asadi@modares.ac.ir}}
\author[2]{\nameemail{Marius Lindauer}{m.lindauer@ai.uni-hannover.de}}

\affil[1]{Division of Computer Science and Software Engineering, M\"alardalen University, Sweden}
\affil[2]{Institute of Artificial Intelligence, Leibniz University Hannover, Germany}
\affil[3]{Department of Electrical Engineering, Tarbiat Modares University, Tehran, Iran}

\hypersetup{%
  pdfauthor={Mohammed Loni, Aditya Mohan}, 
  pdftitle={Activation Function Search for Sparse Neural Networks},
  pdfsubject={Activation Function Search},
  pdfkeywords={Sparse Neural Networks, Activation Functions, HPO}
}

\begin{document}

\maketitle
\input{content/Abstract.tex}
\input{content/Introduction.tex}
\input{content/Related_Work.tex}
\input{content/Preliminaries.tex}
\input{content/Method.tex}
\input{content/Experiments.tex}

\input{content/Conclusion.tex}
\input{content/Broader_Impact.tex}

\newpage

\bibliography{bib/strings, bib/lib, bib/local, bib/proc}


\appendix
\input{content/Appendix.tex}

\end{document}

%% file: Utils/packages.tex
\usepackage{microtype}
\usepackage{graphicx}
\usepackage{booktabs} 

\usepackage{hyperref}
\usepackage{array}
\usepackage{eqparbox}
\usepackage{caption}
\usepackage{tabulary}
\usepackage{mathtools}
\usepackage{pgfplots}
\pgfplotsset{compat = 1.3}
\usepackage{tikz}
\usepackage{soul}
\usetikzlibrary {spy}
\usepackage{multirow}
\usepackage{graphicx}
\usepackage{caption}
\mathchardef\mhyphen="2D

\usepackage{bm}
\usetikzlibrary{spy}
    \def\addlegendimage{\csname pgfplots@addlegendimage\endcsname}

\pgfplotsset{ 
cycle list={%
{draw=black,mark=star,solid},
{draw=black, mark=square,solid}}}

\usepackage{amsmath}

\usepackage{amssymb}
\usepackage{mathtools}

\usepackage{amsthm}

\usepackage[capitalize,noabbrev]{cleveref}

\usepackage{tikz}
\usetikzlibrary{shapes.geometric}
\usetikzlibrary{positioning,shapes,shadows,arrows,calc,fit}
\usetikzlibrary{intersections}
\pgfdeclarelayer{background}
\pgfdeclarelayer{foreground}
\pgfsetlayers{background,main,foreground}

\usepackage[textsize=tiny]{todonotes}

\usepackage{xcolor}
\usepackage{xspace}
\newlist{mylist}{enumerate*}{1}
\setlist[mylist]{label=(\roman*)}

\usepackage[raggedright]{sidecap}
\sidecaptionvpos{figure}{t} 

\definecolor{HOP_Orange}{RGB}{255, 127, 14}
\definecolor{HOP_Blue}{RGB}{31, 119, 180}

\definecolor{my_magenta}{rgb}{0.796875,0.46875,0.734375}
\definecolor{my_green}{rgb}{0.0078125,0.6171875,0.44921875}
\definecolor{my_gold}{rgb}{0.8671875,0.55859375,0.01953125}
\definecolor{my_gray}{rgb}{0.5,0.5,0.5}
\definecolor{my_blue}{rgb}{0.4,0.4,1.}
\definecolor{my_green2}{HTML}{63a76b}
\definecolor{blue}{HTML}{526fae}
\usepackage{bm}

%% file: Utils/commands.tex
\theoremstyle{plain}

\theoremstyle{definition}

\theoremstyle{remark}

%% file: Utils/macros.tex
\newcommand{\ourname}{\texttt{SAFS}\xspace}

\newcommand{\DNNs}{Deep Neural Networks\xspace}

\newcommand{\SNNs}{Sparse Neural Networks\xspace}

\newcommand{\snn}{sparse neural network\xspace}
\newcommand{\snns}{sparse neural networks\xspace}

\newcommand{\af}{activation function\xspace}
\newcommand{\afs}{activation functions\xspace}

%% file: content/Abstract.tex
\begin{abstract}
\label{sec:abstract}

Sparse Neural Networks (SNNs) can potentially demonstrate similar performance to their dense counterparts while saving significant energy and memory at inference. However, the accuracy drop incurred by SNNs, especially at high pruning ratios, can be an issue in critical deployment conditions. While recent works mitigate this issue through sophisticated pruning techniques, we shift our focus to an overlooked factor: hyperparameters and activation functions. Our analyses have shown that the accuracy drop can additionally be attributed to
\begin{mylist}
    \item Using ReLU as the default choice for activation functions unanimously, and
    \item Fine-tuning SNNs with the same hyperparameters as dense counterparts. 
\end{mylist}
Thus, we focus on learning a novel way to tune activation functions for sparse networks and combining these with a separate hyperparameter optimization (HPO) regime for sparse networks. 
By conducting experiments on popular DNN models (LeNet-5, VGG-16, ResNet-18, and EfficientNet-B0) trained on MNIST, CIFAR-10, and ImageNet-16  datasets, we show that the novel combination of these two approaches, dubbed \underline{S}parse \underline{A}ctivation \underline{F}unction \underline{S}earch, short: \ourname, results in up to $15.53\%$, $8.88\%$, and $6.33\%$ absolute improvement in the accuracy for LeNet-5, VGG-16, and ResNet-18 over the default training protocols, especially at high pruning ratios.\footnote{Our code is available at \href{https://github.com/automl/SAFS}{\texttt{github.com/automl/SAFS}}}

\end{abstract}

%% file: content/Introduction.tex
\section{Introduction}
\label{sec:introduction}

\DNNs, while having demonstrated strong performance on a variety of tasks, are computationally expensive to train and deploy. 
When combined with concerns about privacy, energy efficiency, and the lack of stable connectivity, this led to an increased interest in deploying DNNs on resource-constrained devices like micro-controllers and FPGAs \citep{chen-ieee2019}.    

Recent works have tried to address this problem by reducing the enormous memory footprint and power consumption of DNNs. 
These include quantization \citep{zhou-iclr17}, knowledge distillation \citep{hinton-corr15}, low-rank decomposition \citep{jaderberg-bmvc14}, and network sparsification using unstructured pruning (a.k.a. \SNNs) \citep{han-neurips15}. 
Among these, \SNNs (SNNs) have shown considerable benefit through their ability to remove redundant weights \citep{hoefler-jmlr21}. 
However, they suffer from accuracy drop, especially at high pruning ratios; e.g., \cite{mousavi-corr22} report $\approx$54\% reduction in top-1 accuracy for MobileNet-v2 \citep{sandler-cvpr18} trained on ImageNet as compared to non-pruned. 
While significant blame for this accuracy drop goes to sparsification itself, we identified two underexplored, pertinent factors that can additionally impact it:
\begin{mylist}
    \item The \afs of the sparse counterparts are never optimized, with the Rectified Linear Unit (ReLU) \citep{nair-icml10} being the default choice.
    \item The training hyperparameters of the \snns are usually kept the same as their dense counterparts.
\end{mylist}

A natural step, thus, is to understand how the \afs impact the learning process for SNNs. 
Previously, \cite{jaiswal-icml22} and \cite{tessera-corr21} have demonstrated that ReLU reduces the trainability of SNNs since sudden changes in gradients around zero result in blocking gradient flow. Additionally, \cite{apicella-nn20} have shown that a ubiquitous \af cannot prevent typical learning problems such as vanishing gradients. 
While the field of Automated Machine Learning (AutoML) \citep{hutter-book19a} has previously explored optimizing \afs of dense DNNs \citep{ramachandran-iclr18a, loni-micro20, bingham-gecco20}, most of these approaches require a huge amount of computing resources (up to 2000 GPU hours \citep{bingham-gecco20}), resulting in a lack of interest in \af optimization for various deep learning problems. 
On the other hand, attempts to improve the accuracy of SNNs either use sparse architecture search \citep{fedorov-neurips19a, mousavi-corr22} or sparse training regimes \citep{srinivas-ieee17}. 
To our knowledge, there is no efficient approach for optimizing \afs on SNN training. 

\textbf{Paper Contributions:} 
\begin{mylist}
    \item We analyze the impact of \afs and training hyperparameters on the performance of sparse CNN architectures. 
    \item We propose a novel AutoML approach, dubbed \ourname, to tweak the \afs and training hyperparameters of \snns to deviate from the training protocols of their dense counterparts. 
    \item  We demonstrate significant performance gains when applying \ourname with unstructured magnitude pruning to LeNet-5 on the MNIST \citep{lecun-ieee1998} dataset, VGG-16 and ResNet-18 networks trained on the CIFAR-10 \citep{krizhevsky-cifar} dataset, and ResNet-18 and EfficientNet-B0 networks trained on the ImageNet-16 \citep{chrabaszcz-arxiv17a} dataset, when compared against the default training protocols, especially at high levels of sparsity.
\end{mylist}

%% file: content/Related_Work.tex
\section{Related Work}
\label{sec:related-work}

To the best of our knowledge, \ourname is the first automated framework that tweaks the \afs of \snns using a multi-stage optimization method. 
Our study also sheds light on the fact that tweaking the hyperparameters plays a crucial role in the accuracy of \snns. 
Improving the accuracy of \snns has been extensively researched in the past. 
Prior studies are mainly categorized as 
\begin{mylist}
    \item recommending various criteria for selecting insignificant weights,
    \item pruning at initialization or training, and 
    \item optimizing other aspects of sparse networks apart from pruning criteria.
\end{mylist}
In this section, we discuss these methods and compare them with \ourname, and briefly review state-of-the-art research on optimizing \afs of dense networks.   

\subsection{Sparse Neural Network Optimization}
\label{sec:related-work:snn}

\textbf{Pruning Insignificant Weights.} A number of studies have proposed to prune the weight parameters below a fixed threshold, regardless of the training objective \citep{han-neurips15, li-corr16, zhou-neurips19}. 
Recently, \cite{azarian-corr20} and \cite{kusupati-icml20} suggested layer-wise trainable thresholds for determining the optimal value for each layer. 

\noindent\textbf{Pruning at Initialization or Training.} These methods aim to start sparse instead of first pre-training a dense network and then pruning it. 
To determine which weights should remain active at initialization, they use criteria such as using the connection sensitivity \citep{lee-corr18} and conservation of synaptic saliency \citep{tanaka-neurips20}. On the other hand, \cite{mostafa-icml19,  mocanu-nauture18, evci-icml20} proposed to leverage information gathered during the training process to dynamically update the sparsity pattern of kernels.

\noindent\textbf{Miscellaneous Sparse Network Optimization.} \cite{evci-corr19} investigated the loss landscape of \snns and \cite{frankle-icml20} addressed how it is impacted by the noise of Stochastic Gradient Descent (SGD). Finally, \cite{lee-iclr20} studied the effect of weight initialization on the performance of sparse networks. 
While our work also aims to improve the performance of sparse networks and enable them to achieve the same performance as their dense counterparts, we instead focus on the impact of optimizing \afs and hyperparameters of the \snns in a joint HPO setting.

\subsection{Activation Function Search}
\label{sec:related-work:af-search}

Inappropriate selection of \afs results in information loss during forward propagation and the vanishing and/or exploding gradient problems during backpropagation \citep{hayou-icml19}. 
To find the optimal \afs, several studies automatically tuned \afs for dense DNNs, being based on either evolutionary computation \citep{bingham-gecco20, basirat-visigrapp21, nazari-dsd19}, reinforcement learning \citep{ramachandran-iclr18a}, or gradient descent for devising parametric functions \citep{tavakoli-nn21, zamora-aaai22}. 
 
Despite the success of these methods, automated tuning of \afs for dense networks is unreliable for the sparse context since the search spaces for \afs for dense networks are not optimal for sparse networks \citep{dubowski-corr20}. 
The same operations that are successful in dense networks can drastically diminish network gradient flow in sparse networks \citep{tessera-corr21}. 
Additionally, existing methods suffer from significant search costs; e.g., \cite{bingham-gecco20} required 1000 GPU hours per run on NVIDIA\textsuperscript{®} GTX 1080Ti. 
\cite{jin-aaai16} showed the superiority of SReLU over ReLU when training sparse networks as it improves the network's gradient flow. 
However, SReLU requires learning four additional parameters per neuron. 
In the case of deploying networks with millions of hidden units, this can easily lead to considerable computational and memory overhead at inference time. 
\ourname, on the other hand, unifies local search on a meta-level with gradient descent to create a two-tier optimization strategy and obtains superior performance with faster search convergence compared to the state-of-the-art.

%% file: content/Preliminaries.tex
\section{Preliminaries}
\label{sec:preliminaries}
In this section, we develop notations for the later sections by formally introducing the two problems that we address: Network Sparsification and Hyperparameter Optimization. 

\subsection{Network Sparsification}
\label{sec:preliminaries:sparsification}

Network sparsification is an effective technique to improve the efficiency of DNNs for applications with limited computational resources. 
\cite{zhan-corr19} reported that network sparsification could facilitate saving ResNet-18 inference time trained on ImageNet on mobile devices by up to 29.5$\times$. 
Network sparsification generally consists of three stages:
\begin{enumerate}
    \item \textit{Pre-training}: Train a large, over-parameterized model. 
    Given a loss metric $\mathcal{L}_{train}$ and network parameters $\bm{\theta}$, this can be formulated as the task of finding the parameters $\bm{\theta}^{\star}_{pre}$ that minimize $\mathcal{L}_{train}$ on training data $\mathcal{D}_{train}$:

    \begin{equation}\label{Eq:Pruning:Stage_1}
          \bm{\theta}^{\star}_{pre} \in \mathop{\mathrm{argmin}}_{\bm{\theta} \in \bm{\Theta}} \big[\mathcal{L}_{train}(\bm{\theta}; \mathcal{D}_{train})\big]
    \end{equation}
    
    \item \textit{Pruning}: Having trained the dense model, the next step is to remove the low-importance weight tensors of the pre-trained network. 
    This can be done layer-wise, channel-wise, and network-wide. 
    The usual mechanisms either simply set a certain percentage of weights (\emph{pruning ratio}) to zero, or learn a Boolean mask $\bm{m}^\star$ over the weight vector. 
    Both of these notions can be generally captured in a manner similar to the dense training formulation but with a separate loss metric $\mathcal{L}_{prune}$. 
    The objective here is to obtain a pruning mask $\bm{m^\star}$, where $\odot$ represents the masking operation and $N$ represents the size of the mask:

    \begin{eqnarray}\label{Eq:Pruning:Stage_2}
           \bm{m}^\star \in \mathop{\mathrm{argmin}}_{\bm{m}\in\{0,1\}^N} \big[\mathcal{L}_{prune}(\bm{\theta}^\star_{pre} \odot \bm{m}; \mathcal{D}_{train})\big] \,\,\,\, \text{s.t. } \,\,\,\, \|\bm{m}^\star\|_0 \leq  \epsilon 
    \end{eqnarray}

    where $\epsilon$ is a threshold on the minimal number of masked weights.

    \item \textit{Fine-tuning}: The final step is to retrain the pruned network to regain its original accuracy using a fine-tuning \footnote{We use the term fine-tuning interchangeably with re-training} loss $\mathcal{L}_{fine}$, which can either be the same as the training loss or a different kind:

    \begin{equation}\label{Eq:Pruning:Stage_3}
            \bm{\theta}^\star_{fine} \in \mathop{\mathrm{argmin}}_{ \bm{\theta} \in \bm{\Theta}} \big[\mathcal{L}_{fine}(\bm{\theta} ; \bm{\theta}^\star_{pre}\odot \bm{m}^\star, \mathcal{D}_{train})\big]\\
    \end{equation}
    
\end{enumerate}

For the pruning stage, \ourname uses the popular magnitude pruning method \citep{han-neurips15} by removing a certain percentage of weights that have a lower magnitude. 
Compared to structured pruning methods \citep{liu-corr2018}, the magnitude pruning method provides higher flexibility and a better compression rate $\left(\frac{|\bm{\theta}^{\star}_{fine}|}{|\bm{\theta}^\star_{pre}|} \times 100\right)$.
Crucially, \ourname is independent of the pruning algorithm; thus, it can optimize any sparse network. 
  
\subsection{Hyperparameter Optimization (HPO)}
\label{sec:preliminaries:hpo}

We denote the hyperparameter space of the model as $\Lambda$ out of which we sample a hyperparameter configuration $\bm{\lambda} = (\lambda_1, \dots, \lambda_d)$ to be tuned by some HPO methods. 
We assume $c:\bm{\lambda} \to \mathbb{R}$ to be a black-box cost function that maps the selected configuration $\bm{\lambda}$ to a performance metric, such as model-error\footnote{For reasonably sized datasets and models, we estimate this error using k-fold cross-validation.}. 
HPO's goal can then be summarized as the task of finding an optimal configuration $\bm{\lambda}^\star$ minimizing $c$. 
Given the fine-tuned parameters $\bm{\theta}^\star_{fine}$ obtained in \cref{Eq:Pruning:Stage_3}, we define the cost as minimizing a loss $\mathcal{L}_{hp}$ on validation dataset $\mathcal{D}_{val}$ as a bi-level optimization problem:

\begin{eqnarray} \label{Eq:HPO}
    \bm{\lambda}^\star \in  \mathop{\mathrm{argmin}}_{\bm{\lambda} \in \Lambda} c(\bm{\lambda})  =  \mathop{\mathrm{argmin}}_{\bm{\lambda} \in \Lambda} \big[\mathcal{L}_{hp} (\bm{\theta}^\star_{fine}(\bm{\lambda}); \mathcal{D}_{val})\big] \\
    \text{s.t.} \nonumber \\ \bm{\theta}^\star_{fine}(\bm{\lambda}) \in \mathop{\mathrm{argmin}}_{ \bm{\theta} \in \bm{\Theta}} \big[\mathcal{L}_{fine}(\bm{\theta} ; \bm{\theta}^\star_{pre}\odot \bm{m}^\star, \mathcal{D}_{train}, \bm{\lambda})\big] \nonumber
\end{eqnarray}


We note that in principle HPO could also be applied to the training of the original model (\cref{Eq:Pruning:Stage_1}), but we assume that the original is given and we care only about sparsification.

%% file: content/Method.tex
\section{Finding Activation Functions for Sparse Networks}
\label{sec:method}

The aim of \ourname is to find an optimal hyperparameter configuration for pruned networks with a focus on \afs. Given the HPO setup described in \cref{sec:preliminaries:hpo}, we now explain how to formulate the \af search problem and what is needed to solve it.

\subsection{Modelling Activation Functions}
\label{sec:method:problem-definition}

Using optimization techniques requires creating a search space containing promising candidate \afs. Extremely constrained search spaces might not contain novel \afs (\emph{expressivity}) while searching in excessively large search spaces can be difficult (\emph{size}) \citep{ramachandran-iclr18a}. 
Thus, striking a balance between the expressivity and size of the search space is an important challenge in designing search spaces. 

To tackle this issue, we model parametric \afs as a combination of a unary operator $f$ and two learnable scaling factors $\alpha, \beta$. 
Thus, given an input $x$ and output $y$, the \af can be formulated as $y = \alpha f(\beta x)$, which can alternatively be represented as a computation graph shown in \cref{fig:method:activation-a}.

\cref{fig:method:af-variations} illustrates an example of tweaking the $\alpha$ and $\beta$ learnable parameters of the $Swish$ \af. 
We can intuitively see that modifying the suggested learnable parameters for a sample unary operator provides the sparse network additional flexibility to fine-tune \afs \citep{godfrey-smc19, bingham-NN22}. Examples of \afs that we consider in this work have been listed in \cref{appendix:search-details}. 

For sparse networks, this representation allows efficient implementation as well as effective parameterization. 
As we explain further in \cref{sec:method:optimization}, by treating this as a two-stage optimization process, where the search for $f$ is a discrete optimization problem and the search for $\alpha, \beta$ is interleaved with fine-tuning, we are able to make the search process efficient while capturing the essence of input-output scaling and functional transformations prevalent with \afs.
Note that \ourname falls under the category of adaptive activation functions due to introducing trainable parameters \citep{dubey-neurocomputing22}. 
These parameters allow the \afs to smoothly adjust the model with the dataset complexity \citep{zamora-aaai22}.
In contrast to popular adaptive \afs such as PReLU and Swish, \ourname automates \af tuning across a diverse family of \afs for each layer of the network with optimized hyperparameters.

\input{figures/af-variations.tex}

\subsection{Optimization Procedure}
\label{sec:method:optimization}

\ourname performs the optimization layer-wise i.e. we intend to find the activation functions for each layer. 
Given layer indices $i = 1, \dots, L$ of the network of depth $L$ an optimization algorithm needs to be able to select a unary operator $f_i^\star$ and find appropriate scaling factors $(\alpha_i^\star, \beta_i^\star)$. 
We formulate these as two independent objective functions, solved in a two-stage optimization procedure combining discrete and stochastic optimization. \cref{fig:overview} shows an overview of the \ourname pipeline. 

\input{figures/overview.tex}

\paragraph{Stage 1: Unary Operator Search}
The first stage is to find the unary operators after the network has been pruned. 
Crucially, the fine-tuning step happens only after this optimization for the \af has been completed. 
We model the task of finding optimal unary operators for each layer as a discrete optimization problem. Given a pre-defined set of functions $F=\{f_1, f_2, \dots, f_n\}$, we define a space $\mathcal{F}$ of possible sequences of operators $\psi = \langle f_i \mid f_i \in F \rangle_{i\in\{1,\ldots,L\}} \in \mathcal{F}$ of size $L$.  
Our task is to find a sequence $\psi$ after the pruning stage (\cref{Eq:Pruning:Stage_2}). 
Since the pre-trained network parameters $\bm{\theta}^\star_{pre}$ and the pruning mask $\bm{m^\star}$ have already been discovered, we keep them fixed and use them as an initialization point for \af optimization.
The task is formulated as finding the optimal operators given the network parameters, as shown in \cref{eq:method:objective-1}. 
During this step, $\alpha$ and $\beta$ parameters are set to $1$ to focus on the function class first. 

\begin{equation}
\label{eq:method:objective-1}
\psi^\star \in \mathop{\mathrm{argmin}}_{\psi \in \mathcal{F}}\big[\mathcal{L}_{train}(\bm{\theta}^\star_{pre}\odot \bm{m^\star}, \psi; \mathcal{D}_{train})\big]
\end{equation}

\begin{figure}[t]
    \begin{subfigure}[h]{0.4\textwidth}
        \centering
        \includegraphics[width=0.75\textwidth]{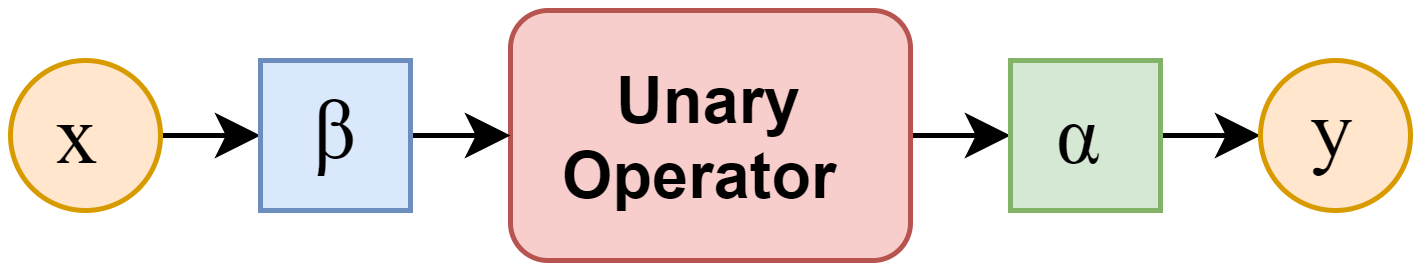}
        \caption{}
        \label{fig:method:activation-a}
    \end{subfigure}%
    \begin{subfigure}[h]{0.45\textwidth}
        \centering
        \includegraphics[width=\textwidth]{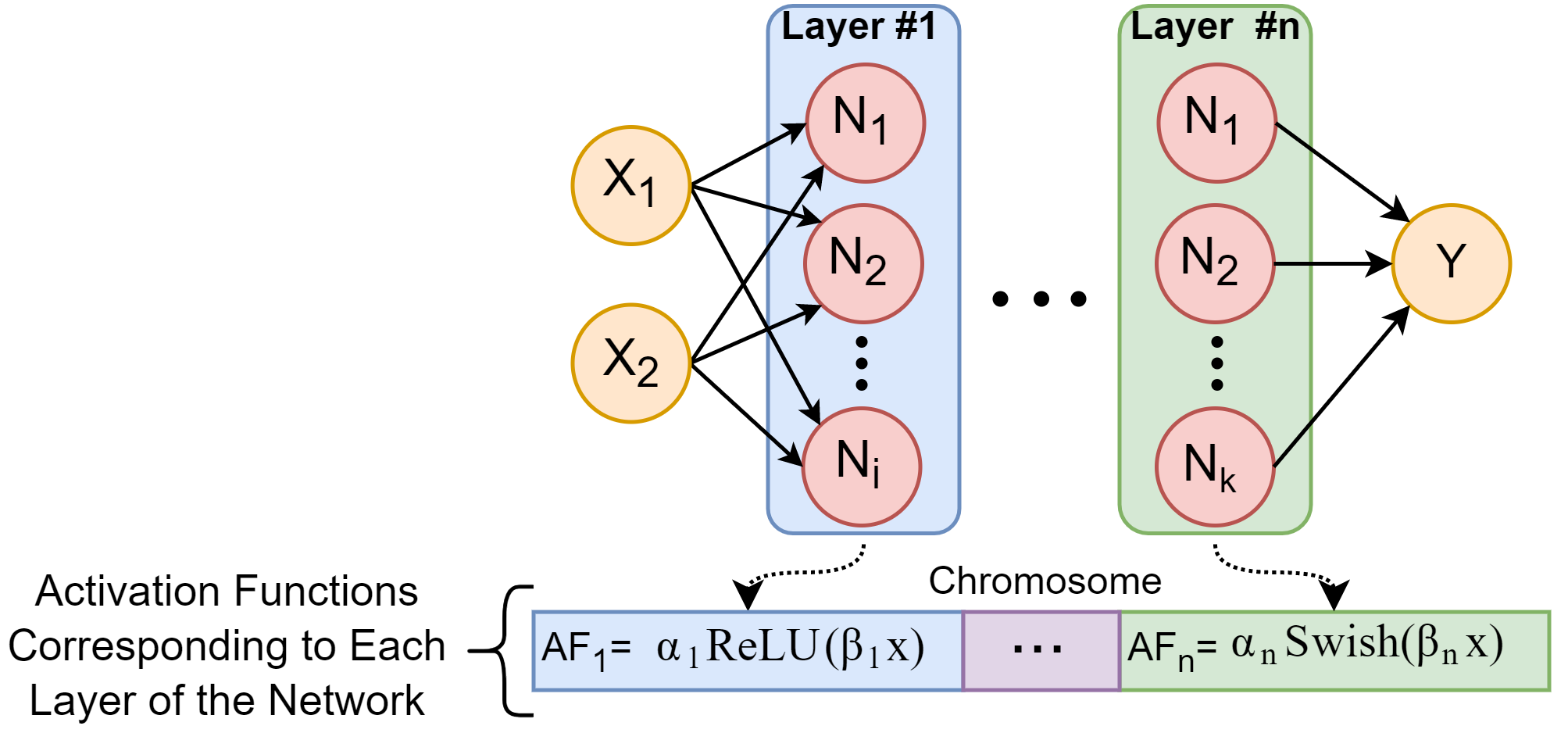}
        \caption{}
        \label{fig:method:activation-b}
    \end{subfigure}
    \caption{(a) \ourname unary activation graph. (b) An example of a solution representing \afs of each layer in the network.}
    \label{Fig:Method:Activation}
\end{figure}

Given the discrete nature of \cref{eq:method:objective-1}, we use Late Acceptance Hill Climbing (LAHC) \citep{burke-ejor17} to iteratively solve it (Please refer to \cref{appendix:search-algorithms} for comparison against other search algorithms). LAHC is a Hill Climbing algorithm that uses a record of the history - \emph{History Length} - of objective values of previously encountered solutions in order to decide whether to accept a new solution. It provides us with two benefits:
\begin{mylist}
    \item Being a semi-local search method, LAHC works on discrete spaces and quickly searches the space to find unary operators.
    \item LAHC extends the vanilla hill-climbing algorithm \citep{selman-encyclopedia06} by allowing worse solutions in the hope of finding a better solution in the future. 
\end{mylist}
We represent the design space of LAHC using a chromosome that is a list of \afs corresponding to each layer of the network. 
\cref{fig:method:activation-b} shows an example of a solution in the design space. The benefit of this representation is its flexibility and simplicity.
For generating a new search candidate (\emph{mutation operation}), we first swap two randomly selected genes from the chromosome, and then, we randomly changed one gene from the chromosome with a new candidate from the list.

\cref{appendix:search-details} lists unary operators considered in this study. 
To avoid instability during training, we ignored periodic operators (e.g., $cos(x)$) and operators containing horizontal ($y=0$) or vertical ($x=0$) asymptotes (e.g., $y=\frac{1}{x}$). 

The process of selecting operators to form the chromosome is repeated for a predefined number of iterations (refer to \cref{appendix:search-details} for the configuration of LAHC).
Given that we have only two mutations per each search iteration, the entire chromosome is not significantly affected. Based on trial runs, we determined a budget of 20 search iterations to provide decent improvement alongside reducing the search cost. 
Each iteration consists of training the network using the selection \afs and measuring the training loss $\mathcal{L}_{train}$ as a fitness metric that needs to be minimized.

A downside of this process is the need to retrain the network for each search iteration, which can be intensive in time and compute resources. 
We circumvent this issue by leveraging a lower fidelity estimation of the final performance. 
Given that the network performance does not vary after a certain number of epochs, we leverage the work by \cite{loni-micro20} and only train the network up to a certain point after which the performance should remain stable.

\textbf{Stage 2: Scaling Factor and HPO} 
Given a learned sequence of optimal operators $\psi$, the next step is to find a sequence $\psi' = \langle (\alpha_i, \beta_i) \mid \alpha_i, \beta_i \in \mathbb{R} \rangle_{i\in \{1,\ldots,L\}}$ representing the scaling factors for each layer. 
We perform this process jointly with the fine-tuning stage (\cref{Eq:Pruning:Stage_3}) and HPO to discover the fine-tuning parameters $\bm{\theta}^\star_{fine}$ and hyperparameters $\bm{\lambda}^\star$ as shown in \cref{eq:method:objective-2}. 

\begin{align}\label{eq:method:objective-2}
\small
    \bm{\lambda}^\star \in  \mathop{\mathrm{argmin}}_{\bm{\lambda} \in \Lambda} c(\bm{\lambda}; \mathcal{D}_{val} ) \,\,\,\, \text{s.t.} \,\,\,\, 
    \psi'^\star, \bm{\theta}^\star_{fine} (\bm{\lambda}) \in \mathop{\mathrm{argmin}}_{\theta \in \bm{\Theta}, \psi' \in \mathbb{R}^{(2,L)}} \bigg [\mathcal{L}_{fine} \big ( (\bm{\theta} \mid \bm{\theta}^\star_{pre}\odot \bm{m}^\star), \psi'; 
    \psi, \mathcal{D}_{train} \big ) \bigg ]  
\end{align}

\noindent
Due to the continuous nature of this stage, we use the Stochastic Gradient Descent (SGD) for solving \cref{eq:method:objective-2}, and use the validation accuracy as a fitness metric for the hyperparameter configuration. 

Treating the scaling factors as learnable parameters allows us to learn them during the fine-tuning state. Thus, the inner optimization in this step has nearly no overhead costs. 
The only additional cost is that of HPO, which we demonstrate in our experiments to be important and worth it since the hyperparameters from training the original model might not be optimal for fine-tuning.

%% file: figures/af-variations.tex
\begin{table}[t]
\captionsetup{justification=centering}
\begin{center}
\begin{tabular}{cc}
\begin{tikzpicture}[scale=0.5]
      
      \begin {axis}[
        title=$\alpha$ Variations,
        style={at={(0.5,1.07)},anchor=north,yshift=-0.1},
        xmin=-3, xmax=3,
        ymin=-1, ymax=3,
        grid=major, 
        grid style={dashed, gray!40},
        xlabel= Input, 
        ylabel= Output,
        legend style={at={(0.5,-0.2)},anchor=north, nodes={scale=1, transform shape}}, 
        legend pos=north west,
        label style={font=\large}, tick label style={font=\large} 
        ]
        
\addplot[olive, thick, domain = -3:3] {(0.1 * x)/(1 + exp(-x)) };
\addplot[red, thick, domain = -3:3] {(0.5 * x)/(1 + exp(-x)) };
\addplot[blue, thick, domain = -3:3] {(1.0 * x)/(1 + exp(-x)) };
\addplot[teal, thick, domain = -3:3] {(1.5 * x)/(1 + exp(-x)) };
\addplot[violet, thick, domain = -3:3] {(2 * x)/(1 + exp(-x)) };

\legend{$\alpha=0.1$, $\alpha=0.5$, $\alpha=1.0$, $\alpha=1.5$, $\alpha=2.0$}
        
      \end{axis}
    \end{tikzpicture}
    &
\begin{tikzpicture}[scale=0.5]
      \begin {axis}[
        title=$\beta$ Variations,
        style={at={(0.5,1.07)},anchor=north,yshift=-0.1},
        xmin=-3, xmax=3,
        ymin=-2, ymax=3,
        grid=major, 
        grid style={dashed, gray!40},
        xlabel= Input, 
        ylabel= Output,
        legend style={at={(0.5,-0.2)},anchor=north, nodes={scale=1, transform shape}}, 
        legend pos=north west,
        label style={font=\large}, tick label style={font=\large} 
        ]
        
\addplot[olive, thick, domain = -3:3] {(x)/(1 + exp(-0.1 * x)) };
\addplot[red, thick, domain = -3:3] {(x)/(1 + exp(-0.5 * x)) };
\addplot[blue, thick, domain = -3:3] {(x)/(1 + exp(-1.0 * x)) };
\addplot[teal, thick, domain = -3:3] {(x)/(1 + exp(-1.5 * x)) };
\addplot[violet, thick, domain = -3:3] {(x)/(1 + exp(-2.0 * x)) };

\legend{$\beta=0.1$, $\beta=0.5$, $\beta=1.0$, $\beta=1.5$, $\beta=2.0$}
        
      \end{axis}
    \end{tikzpicture}
\\
\textbf{(a) } & \textbf{(b)}  
\end{tabular}
\captionof{figure}{Modifying (a) $\alpha$ and (b) $\beta$ learnable scaling factors of the $Swish$ \af.}
\vspace{-17pt}
\label{fig:method:af-variations}
\end{center}
\end{table}

%% file: figures/overview.tex
\begin{figure}[htbp]
    \centering
    \tikzstyle{package}=[rectangle, draw=black, text centered, fill=white, drop shadow]
    \tikzstyle{loss}=[circle, draw=black, text centered, fill=white]
    \tikzstyle{myarrow}=[->, thick]
\scalebox{0.7}{
    
    \begin{tikzpicture}[node distance=3.8em,
  title/.style={font=\scriptsize\color{black!50}\ttfamily},
  typetag/.style={rectangle, draw=black!50, font=\scriptsize\ttfamily, anchor=south}]

    \node (train)[package] {$\mathcal{D}_{\text{train}}$};
    \node (trainloss)[loss, right of=train] {$\mathcal{L}_{\text{train}}$};
    \node (trainweights)[package, right of=trainloss] {$\bm{\theta}^\star_{\text{pre}}$};
    \node (pruneloss)[loss, right of=trainweights] {$\mathcal{L}_{\text{prune}}$};
    \node (prunemask)[package, right of=pruneloss] {$\bm{m}^\star$};
    \node (trainloss_two)[loss, right of=prunemask] {$\mathcal{L}_{\text{train}}$};
    \node (operators)[package, right of=trainloss_two] {$\psi^\star$};
    \node (fineloss)[loss, right of=operators] {$\mathcal{L}_{\text{fine}}$};
    \node (fineweights)[package, right of=fineloss, node distance=4.2em] {$\bm{\theta}^\star_{\text{fine}}, \psi'^\star$};

    \node (hploss)[loss, above of=fineloss] {$\mathcal{L}_{\text{hp}}$};
    \node (hp)[package, right of=hploss] {$\bm{\lambda}^\star$};
    \node (val)[package, left of=hploss] {$\mathcal{D}_{\text{val}}$};

    \draw[->, thick] (train) to (trainloss);
    \draw[->, thick] (trainloss) to (trainweights);
    \draw[->, thick] (trainweights) to (pruneloss);
    \draw[->, thick] (pruneloss) to (prunemask);
    \draw[->, thick] (prunemask) to (trainloss_two);
    \draw[->, thick] (trainloss_two) to (operators);
    \draw[->, thick] (operators) to (fineloss);
    \draw[->, thick] (fineloss) to (fineweights);

    \draw[-, thick] (train.south) |- ($(pruneloss.south)+(0,-0.25)$);
    \draw[-, thick] (train.south) |- ($(fineloss.south)+(0,-0.345)$);

    \draw[->, thick] ($(fineloss.south)+(0,-0.345)$) -- (fineloss.south);
    \draw[->, thick] ($(pruneloss.south)+(0,-0.25)$) -- (pruneloss.south);

    \draw[->, thick] (val) -- (hploss);
    \draw[->, thick] (hploss) -- (hp);
    \draw[<->, thick] (hploss) -- (fineloss);

    \end{tikzpicture}
    }
    \caption{Overview of the entire \ourname pipeline.}
    \label{fig:overview}
\end{figure}

%% file: content/Experiments.tex
\section{Experiments}
\label{sec:experiments}

We categorize the experiments based on the research questions this work aims to answer. \cref{sec:experiments:experimental-setup} introduces the experimental setup. \cref{sec:experiments:af-optimization-impact} motives the problem of tuning activation functions for SNNs. \cref{sec:experiments:snn-training} introduces the need for HPO with activation tuning for SNNs. In \cref{sec:experiments:vanilla}, we compare \ourname against different baselines. \cref{appendix:tradeoff} provides an accuracy improvement vs. compression ratio trade-off to compare \ourname with state-of-the-art network compression methods. In \cref{sec:experiments:pruning-ratio} we compare the performance of \ourname for various pruning ratios. In \cref{sec:experiments:insights} we provide insights on the activation functions learned by \ourname.
Finally, we ablate \ourname in \cref{sec:experiments:ablation} to determine the impact of different design choices.

\subsection{Experimental Setup}
\label{sec:experiments:experimental-setup}

\textbf{Datasets.} To evaluate \ourname, we use MNIST \citep{lecun-ieee1998}, CIFAR-10 \citep{krizhevsky-cifar} and ImageNet-16 \citep{chrabaszcz-arxiv17a} public classification datasets. Note that ImageNet-16 includes all images of the original ImageNet dataset, resized to 16$\times$16 pixels. All HPO experiments were conducted using SMAC3 \citep{lindauer-jmlr22a}. \cref{appendix:search-details} presents the rest of the experimental setup. 

\subsection{The Impact of Tweaking Activation Functions on the Accuracy of SNNs}
\label{sec:experiments:af-optimization-impact}

To validate the assumption that \afs indeed impact the accuracy, we investigated whether \afs currently used for dense networks \citep{evci-aaai22} are still reliable in the sparse context. \cref{fig:motivation:AF} shows the impact of five different \afs on the accuracy of sparse architectures with various pruning ratios. To measure the performance during the search stage, we use a three-fold validation approach. However, we report the test accuracy of \ourname to compare our results with other baselines.

Our conclusions from this experiment can be summarised as follows:
\begin{mylist}
    \item ReLU does not perform the best in all scenarios. We see that SRS, Swish, Tanh, Symlog, FLAU, and PReLU outperform ReLU on higher sparsity levels. 
    Thus, the decision to use ReLU unanimously can limit the potential gain in accuracy.
    
    \item As we increase the pruning ratio to 99\% (extremely sparse networks), despite the general drop in accuracy, the difference in the sparse and dense networks' accuracies vary greatly depending on the activation function. 
    Thus, the choice of activation function for highly sparse networks becomes an important parameter.
\end{mylist}
We need to mention that despite the success of \ourname in providing higher accuracy, it needs 47 GPU hours in total for learning \afs and optimal HPs. On the other hand, refining a \snn takes $\approx$ 3.9 GPU hours.

\subsection{The Difficulty of Training \SNNs}
\label{sec:experiments:snn-training}

\begin{figure}[htbp]
    \begin{subfigure}[h]{0.5\textwidth}
        \centering
        \includegraphics[width=0.95\textwidth]{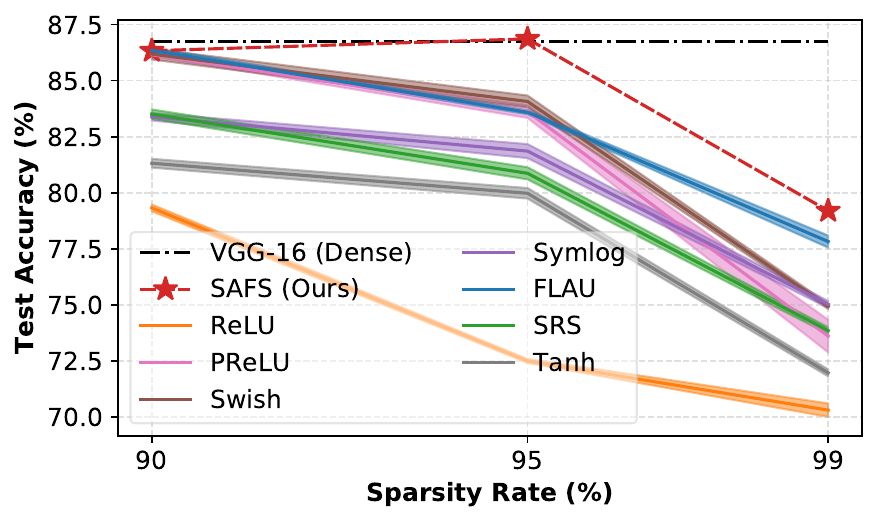}
        \caption{} 
        \label{fig:motivation:AF}
    \end{subfigure}%
    \begin{subfigure}[h]{0.5\textwidth}
        \centering
        \includegraphics[width=0.95\textwidth]{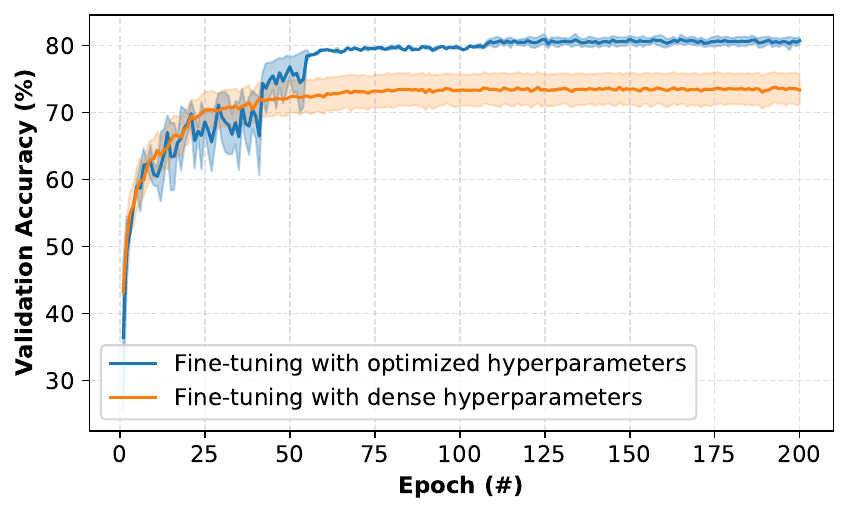}
        \caption{} 
        \label{fig:motivation:HPO}  
    \end{subfigure}
\caption{(a) CIFAR-10 test accuracy on sparse VGG-16 with various \afs customized for dense networks with a 3-fold cross-validation procedure. The bold line represents the mean across the folds, while the shaded area represents the Confidence Intervals across the folds. (b) Fine-tuning sparse VGG-16 on CIFAR-10 with different training hyperparameters with three different random seeds. The pruning ratio is 99\%. As shown, fine-tuning with dense hyperparameters results in inefficient training of SNNs.}
\end{figure}

Currently, most algorithms for training sparse DNNs use configurations customized for their dense counterparts, e.g., starting from a fixed learning scheduler. 
To validate the need for optimizing the training hyperparameters of the sparse networks, we used the dense configurations as a baseline against hyperparameters learned by an HPO method. 
\cref{fig:motivation:HPO} shows the curves of fine-tuning sparsified VGG-16 with 99\% pruning ratio trained on CIFAR-10. 
The training has been performed with the hyperparameters of the dense network (\textcolor{HOP_Blue}{Blue}), and training hyperparameters optimized using SMAC3 (\textcolor{HOP_Orange}{Orange}).

We optimized the learning rate, learning rate scheduler, and optimizer hyperparameters with the range specified in \cref{appendix:search-details} (\cref{tab:search_details}). 
The type and range of hyperparameters are selected based on recommended ranges from deep learning literature \citep{simonyan-arxiv14a, subramanian-bigdata22}, SMAC3 documentation \citep{lindauer-jmlr22a}, and from the various open-source libraries\footnote{\url{https://www.kaggle.com/datasets/keras/vgg16/code}} used to implement VGG-16. 
To prevent overfitting on the test data, we optimized the hyperparameters on validation data and tested the final performance on the test data. 
The poor performance ($7.17\%$ accuracy reduction) of the SNN learning strategy using dense parameters motivates the need for a separate sparsity-aware HPO regime.

\subsection{Comparison with Magnitude Pruning Baselines}
\label{sec:experiments:vanilla}

\cref{tab:Experiments:Comparison} shows the results of optimizing sparse VGG-16 \afs trained on CIFAR-10 using \ourname with 99\% pruning ratio. An average of three runs has been reported. Results show that \ourname provides $8.88\%$ absolute accuracy improvement for VGG-16 and $6.33\%$ for ResNet-18 trained on CIFAR-10 when compared against a vanilla magnitude pruning baseline. 
\ourname additionally yields $1.8\%$ absolute Top-1 accuracy improvement for ResNet-18 and $1.54\%$ for EfficientNet-B0 trained on ImageNet-16 when compared against a vanilla magnitude pruning baseline.
SReLU \citep{jin-aaai16} is a piece-wise linear \af that is formulated by four learnable parameters. 
\cite{mocanu-nauture18, curci-corr21, tessera-corr21} have shown SReLU performs excellently for \snns due to improving the network's gradient flow.
Results show that \ourname provides $15.99\%$ and $19.17\%$ higher accuracy compared to training VGG-16 and ResNet-18 with SReLU \af on CIFAR-10. 
Plus, \ourname provides $0.88\%$ and $1.28\%$ better accuracy compared to training ResNet-18 and EfficientNet-B0 with SReLU \af on the ImageNet-16 dataset.
Lastly, \cref{appendix:gradient-flow} shows that  \ourname significantly improves the gradient flow of \snns, which is associated with optimized activation functions and efficient training protocol.

\begin{table}[htbp]
\centering
\caption{Refining \snn \afs with different methods. }
\label{tab:Experiments:Comparison}
\resizebox{\textwidth}{!}{%
\begin{tabular}{c|cc|cc}
\hline
\textbf{Magnitude Pruning} & \multicolumn{2}{c}{\textbf{CIFAR-10 (Top-1)}}     & \multicolumn{2}{c}{\textbf{ImageNet-16$^\ddagger$ (Top-1 / Top-5)}}                   \\  \cline{2-5}   
\textbf{\citep{han-neurips15}} & VGG-16 & ResNet-18 & ResNet-18 & EfficientNet-B0\\ \hline
Original Model  (Dense) & 86.76\% &  89.86\% &  25.42\% / 47.26\% & 18.41\% / 37.45\% \\
Vanilla Pruning (\underline{Baseline}) & 70.32\%  &  77.55\%  & 11.32\% / 25.59\%  & 10.96\% / 25.62\%\\
SReLU &  63.21\% & 64.71\% & 12.24\% / 26.89\% & 11.22\% / 25.98\% \\
\ourname (Ours)   & 79.2\% \textbf{\color{blue}(+8.88\%)} & 83.88\% \textbf{\color{blue}(+6.33\%)} & 13.12\% \textbf{\color{blue}(+1.8\%)} / 28.94\%   & 12.5\% \textbf{\color{blue}(+1.54\%)} / 27.15\%  \\ \hline
\multicolumn{5}{l}{$^\ddagger$ The Top-1 accuracy of WideResNet-20-1 on ImageNet-16 is 14.82\% \citep{chrabaszcz-arxiv17a}.} \\ \hline
\end{tabular}
}    
\end{table}

\subsection{Evaluation of \ourname with Various Pruning Ratios}
\label{sec:experiments:pruning-ratio}

\cref{fig:motivation:AF} compares the performance of VGG-16 fine-tuned by \ourname and the default training protocol on CIFAR-10 over three different pruning ratios including $90\%$, $95\%$, and $99\%$. 
Results show that \ourname is extremely effective by achieving $1.65\%$, $7.45\%$, and $8.88\%$ higher accuracies compared to VGG-16 with ReLU \afs fine-tuned with the default training protocol at $90\%$, $95\%$, and $99\%$ pruning ratios. Plus, \ourname is better than \afs designed for dense networks, especially for networks with a 99\% pruning ratio.

\subsection{Insights on Searching for Activation Functions}
\label{sec:experiments:insights}

\begin{figure}[htbp]
\centering
\includegraphics[width=0.65\textwidth]{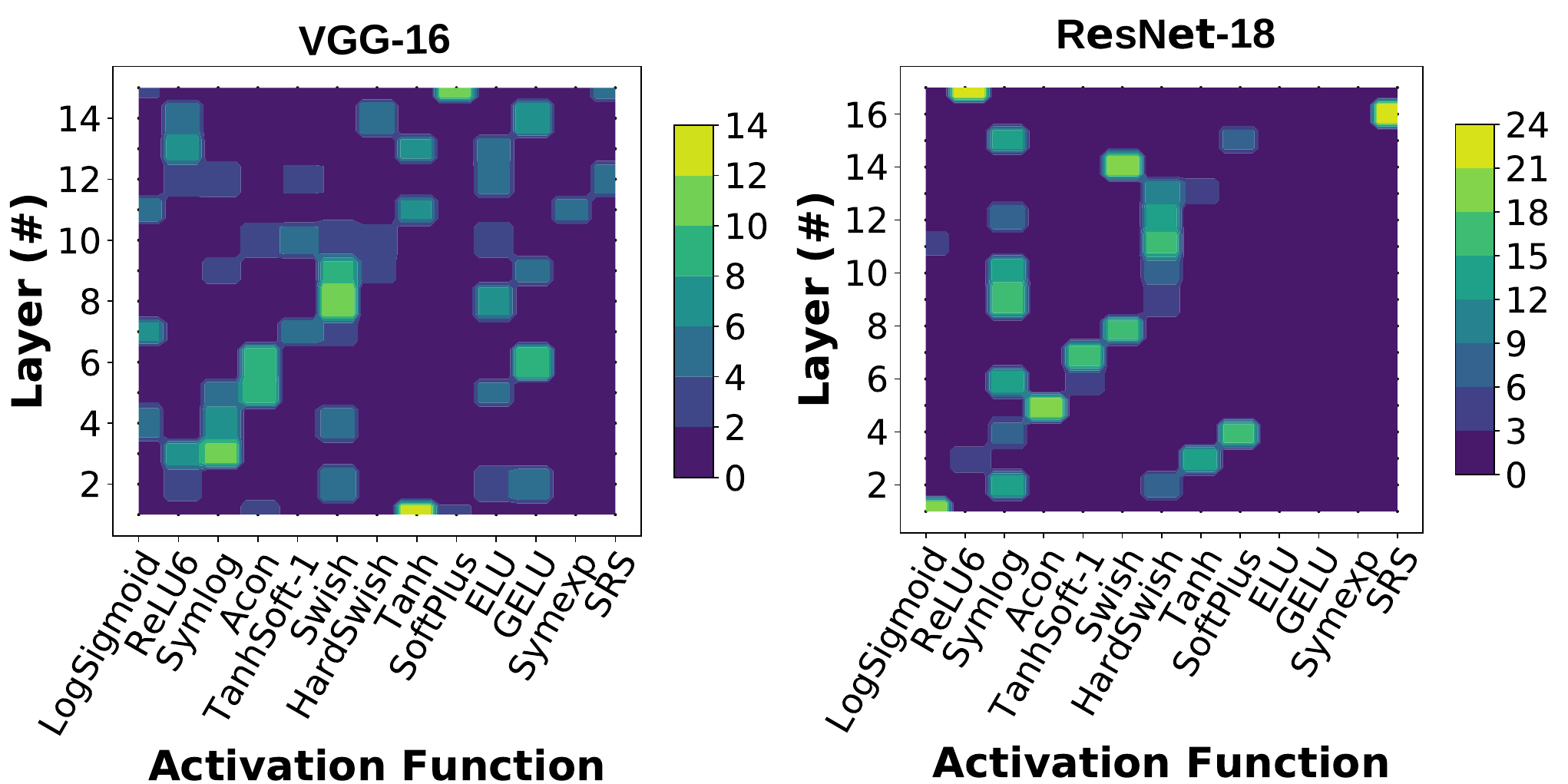}
\caption{Frequency of Occurring unary operator in the first learning stage ($\alpha=\beta=1$) for VGG-16 and ResNet-18 trained on CIFAR-10 with 99\% pruning ratio. }
\label{fig:Insights}
\end{figure}

\cref{fig:Insights} presents the dominance pattern of each unary operator in the first learning stage ($\alpha=\beta=1$) for the CIFAR-10 dataset.
The results are the average of three runs with different random seeds. The unit of the color bar is the number of seeing a specific \af across all search iterations for the first learning stage. According to the results, it is evident that
\begin{mylist}
    \item Symexp and ELU are unfavorable activation functions, 
    \item Symlog and Acon are dominant \afs while being used in the early layers, and
    \item Overall Swish and HardSwish are good, but they mostly appear in the middle layers. 
\end{mylist}

\subsection{Ablation Study}
\label{sec:experiments:ablation}
We study the effect of each individual optimization stage of \ourname on the performance of sparse VGG-16 and ResNet-18 trained on CIFAR-10 in \cref{tab:ablation:af-hpo}. 
Results show that each individual contribution provides higher accuracy for both VGG-16 and ResNet-18. However, the maximum performance is attained by \ourname ($+15.53\%$, $+8.88\%$, $+6.33\%$, and $+1.54\%$ for LeNet-5, VGG-16, ResNet-18, and EfficientNet-B0), where we first learn the most accurate unary operator for each layer and then fine-tune scaling factors with optimized hyperparameters. 

\begin{table}[htbp]
        \centering
        \caption{Ablation Study on optimizing \afs of SNNs with 99\% pruning ratio.}
        \label{tab:ablation:af-hpo}
        \resizebox{\linewidth}{!}{
        \begin{tabular}{ccc|ccc}
        \hline
        \textbf{CNN} &\textbf{Dense}  &  \textbf{Magnitude} &  \multicolumn{3}{c}{\textbf{Learning Activation Functions$^\mp$ }}  
        \\ \cline{4-6}
        \textbf{Model$^\star$} &\textbf{Model}  & \textbf{Pruning} & (Stage 1)$^\dagger$ &  (Stage 2)$^\ddagger$ & \ourname (Stage 1 + Stage 2) \\
        \hline
        LeNet-5 & 98.49\%  & 46.69\% & 61.63\% &  60.2\% & 62.22\% \textbf{\color{blue}(+15.53\%)}  \\
        VGG-16 & 86.76\%  & 70.32\% & 78.11\% &  80.97\% & 79.2\% \textbf{\color{blue}(+8.88\%)}  \\
        ResNet-18 & 89.86\%  & 77.55\% & 79.34\% & 82.74\% &  83.88\% \textbf{\color{blue}(+6.33\%)}  \\
        EfficientNet-B0 & 18.41\%  & 10.96\% & 11.84\% & 11.7\% &  12.5\% \textbf{\color{blue}(+1.54\%)}  \\
        \hline
\multicolumn{6}{l}{$^\star$ Lenet-5, VGG-16, ResNet-18, and EfficientNet-B0 are trained on MNIST, CIFAR-10, CIFAR-10, and ImageNet-16, respectively.}\\
\multicolumn{6}{l}{$^\mp$ ReLU is the default \af for Lenet-5, VGG-16, and ResNet-18. Swish is the default \af for EfficientNet-B0.}\\
\multicolumn{6}{l}{$^\dagger$ Learning \afs by only using the first stage of \ourname ($\alpha=\beta=1$ and without using HPO).}\\
\multicolumn{6}{l}{$^\ddagger$ Learning $\alpha$ and $\beta$ for the ReLU operator with optimized hyperparameters.} \\
        \hline
        \end{tabular}
        }
\end{table}

%% file: content/Conclusion.tex
\section{Conclusion}
\label{sec:conclusion}

In this paper, we studied the impact of activation functions on training \snns and use this to learn new \afs. To this end, we demonstrated that the accuracy drop incurred by training SNNs uniformly with ReLU for all units can be partially mitigated by a layer-wise search for \afs. We proposed a novel two-stage optimization pipeline that combines discrete and stochastic optimization to select a sequence of \afs for each layer of an SNN, along with discovering the optimal hyperparameters for fine-tuning. Our method \ourname provides significant improvement by achieving up to $8.88\%$ and $6.33\%$ higher accuracy for VGG-16 and ResNet-18 on CIFAR-10 over the default training protocols, especially at high pruning ratios. Crucially, since \ourname is independent of the pruning algorithm, it can optimize any sparse network.

%% file: content/Broader_Impact.tex
\section{Limitations and Broader Impact}
\label{Sec:Broader_Impact}

\textbf{Broader Impact.} The authors have determined that this work will have no negative impacts on society or the environment, since this work does not address any concrete application.

\noindent\textbf{Future Work and Limitations.} Sparse Neural Networks (SNNs) enable the deployment of large models on resource-limited devices by saving computational costs and memory consumption. In addition, this becomes important in view of decreasing the carbon footprint and resource usage of DNNs at inference time. We believe this opens up new avenues of research into methods that can improve the accuracy of SNNs. We hope that our work motivates engineers to use SNNs more than before in real-world products as \ourname provides SNNs with similar performance to dense counterparts. Some immediate directions for extending our work are 
\begin{mylist}
    \item leveraging the idea of accuracy predictors \citep{li-corr23} in order to expedite the search procedure. 
    \item SNNs have recently shown promise in application to techniques for sequential decision-making problems such as Reinforcement Learning \citep{vischer-iclr22, graesser-icml22}. We believe incorporating \ourname into such scenarios can help with the deployability of such pipelines.
\end{mylist}

\ourname has been evaluated on diverse datasets, including MNIST, CIFAR-10, and ImageNet-16, and various network architectures such as LeNet-5, VGG-16, ResNet-18, and EfficientNet-B0. 
While the current results demonstrate the general applicability of our method and signs of scalability, we believe further experiments on larger datasets and more scalable networks would be an interesting avenue for future work. 

\section*{Acknowledgements}
Aditya Mohan and Marius Lindauer were supported by the German Federal Ministry of the Environment, Nature Conservation, Nuclear Safety and Consumer Protection (GreenAutoML4FAS project no. 67KI32007A). Mohammad Loni was supported by the HiPEAC project, a European Union’s Horizon 2020 research and innovation program under grant agreement number 871174.

%% file: content/Appendix.tex
\newpage

\section{Evaluation of Various Search Algorithms}
\label{appendix:search-algorithms}

\cref{fig:various-search-algorithms} shows the trend of search performance for finding the best unary operators (\cref{eq:method:objective-1}) over popular search algorithms, including Late-Acceptance-Hill-Climbing (LAHC), Simulated Annealing (SA), Random Search (RS), and Bayesian Optimization (BO). VGG-16 is trained on CIFAR-10 with a 99\% pruning ratio. The bold line represents the mean across three random seeds, while the shaded area represents the confidence intervals. Overall, the observation is that \ourname's search algorithm, LAHC, finds better \afs  than other counterparts with an equal search budget.

\begin{table}[hbpt]
\centering

\resizebox{\textwidth}{!}{

\begin{tabular}{cc}
\includegraphics[width=\textwidth]{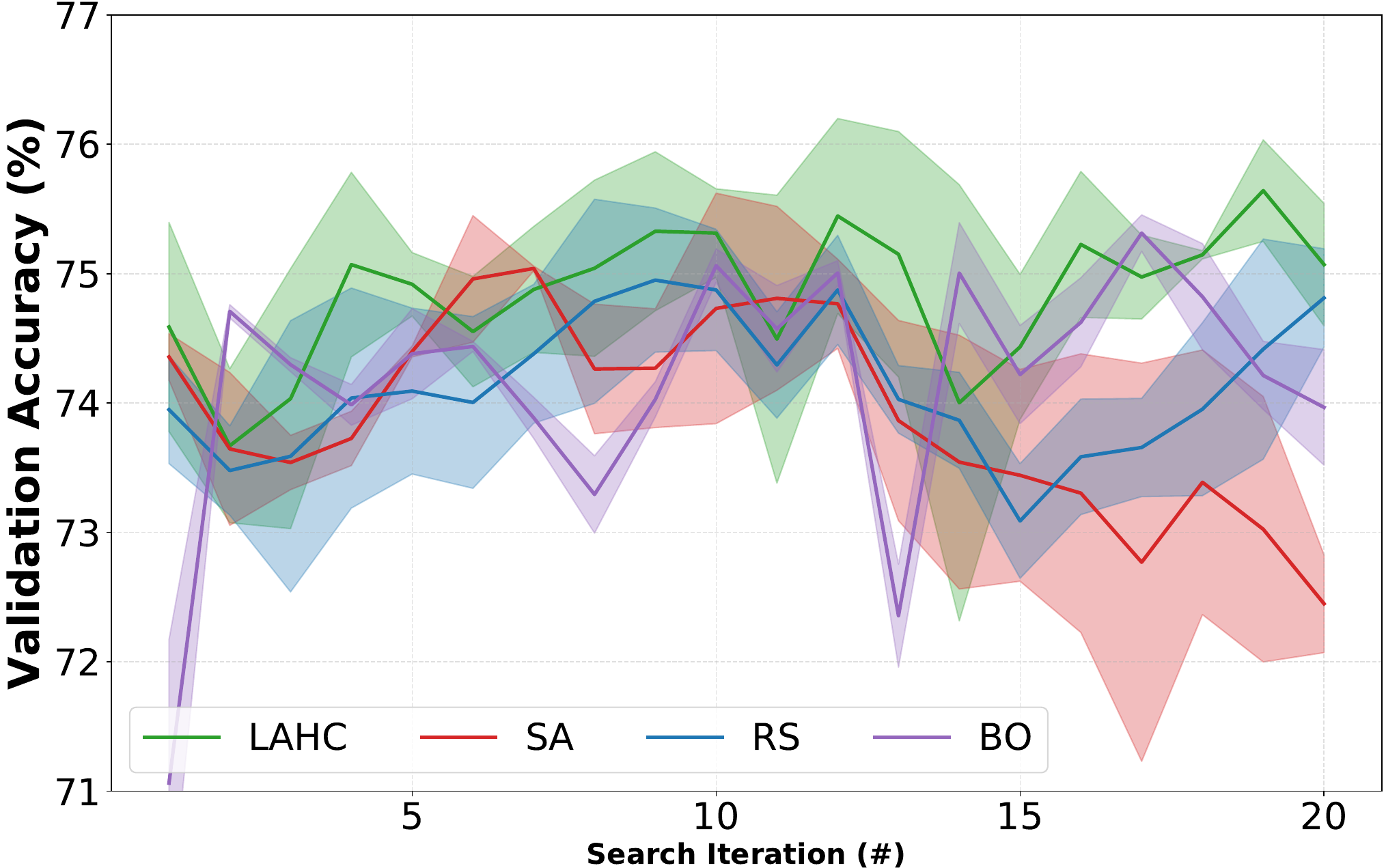}
&
\includegraphics[width=\textwidth]{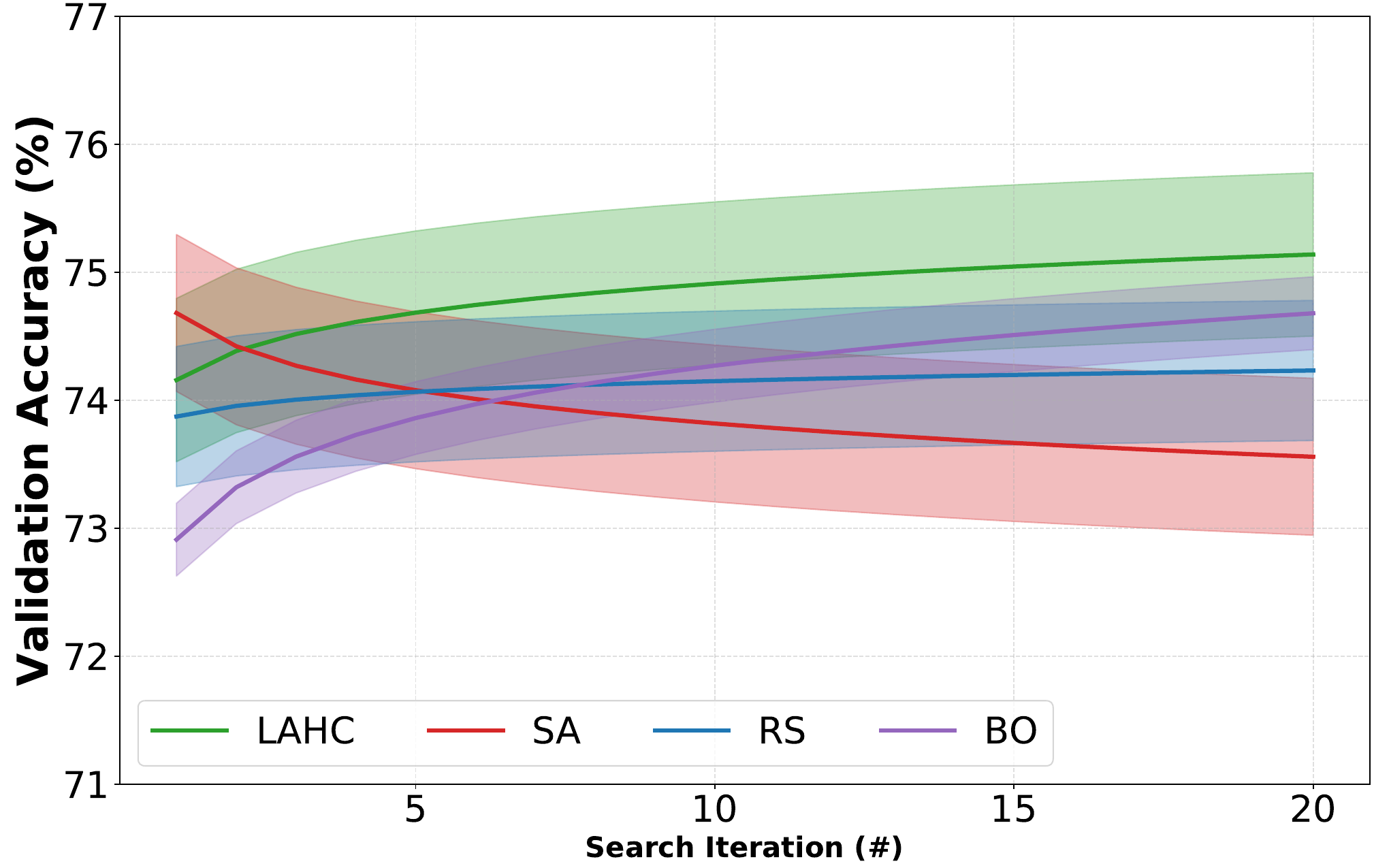}
\\
\textbf{\huge (a) Plotting Search Objective (Vali. Acc.) per Iteration } &  \textbf{\huge (b) Search Trend}
\end{tabular}
 }
\captionof{figure}{Comparison of different search algorithms (LAHC, SA, RS, BO) for finding the best unary operators for sparse VGG-16 with 99\% pruning ratio trained on CIFAR-10. The bold line represents the mean across three random seeds, while the shaded area represents the confidence intervals. (a) Showing raw data. (b) Using a smoothing average function (logarithmic) for representing  the trend of data.} 
\label{fig:various-search-algorithms}
\end{table}

\section{Comparing Gradient Flow of \ourname with the Vanilla Pruning}
\label{appendix:gradient-flow}

\cref{fig:gradient-flow} compares the gradient flow of the sparse VGG-16 trained on CIFAR-10 using \ourname (\textcolor{HOP_Blue}{Blue} and the vanilla pruning (\textcolor{HOP_Orange}{Orange}). As a reminder, gradient flow is the first-order approximation of the decrease in loss after each gradient step, thus the higher the value is the better. Results show that \ourname significantly improves this metric, which is associated with optimized \afs and efficient training of \snns.

\begin{table}[hbpt]
\centering
\resizebox{\textwidth}{!}{
\begin{tabular}{cc}
\includegraphics[width=0.2\textwidth]{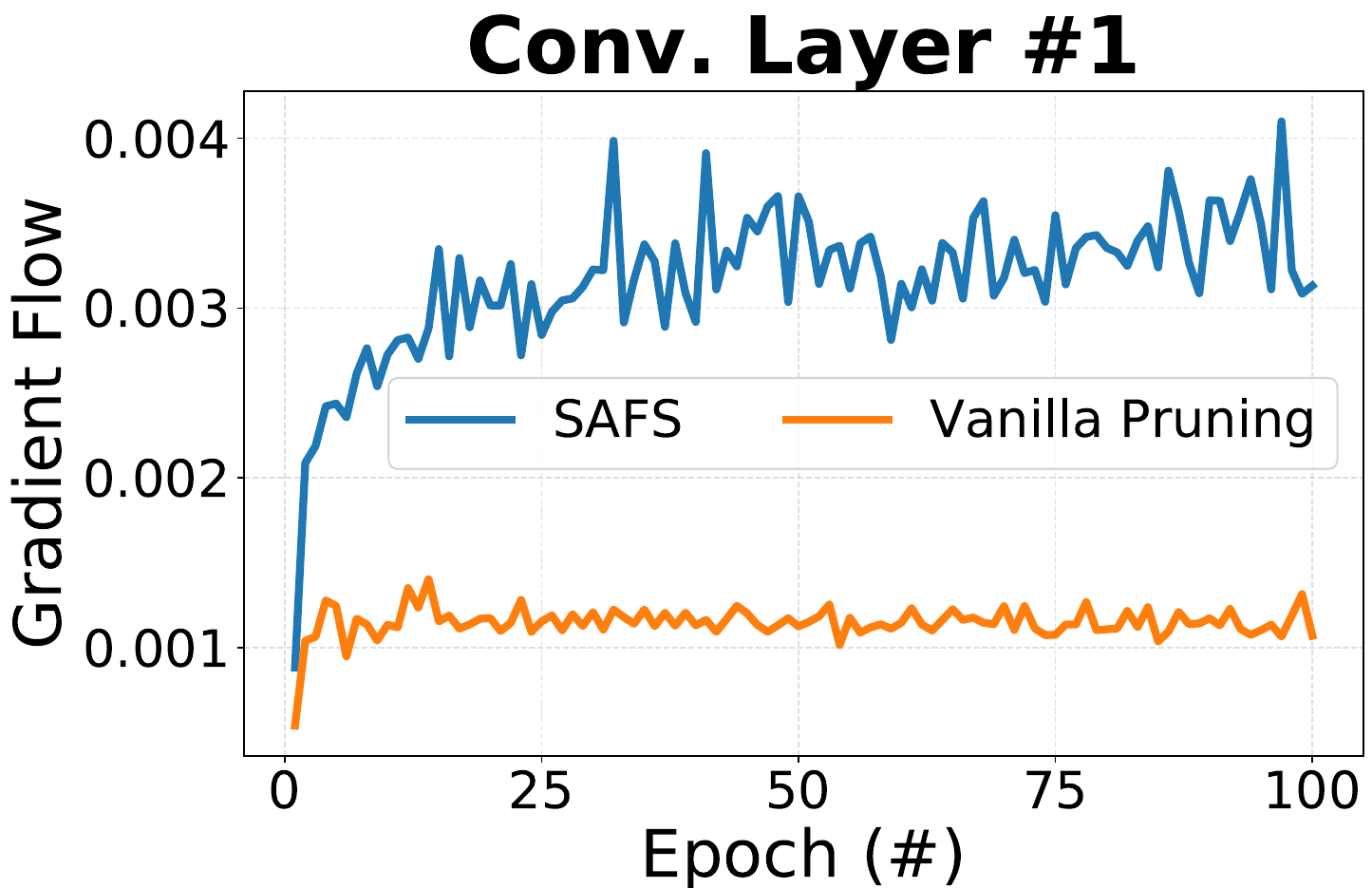}
&
\includegraphics[width=0.2\textwidth]{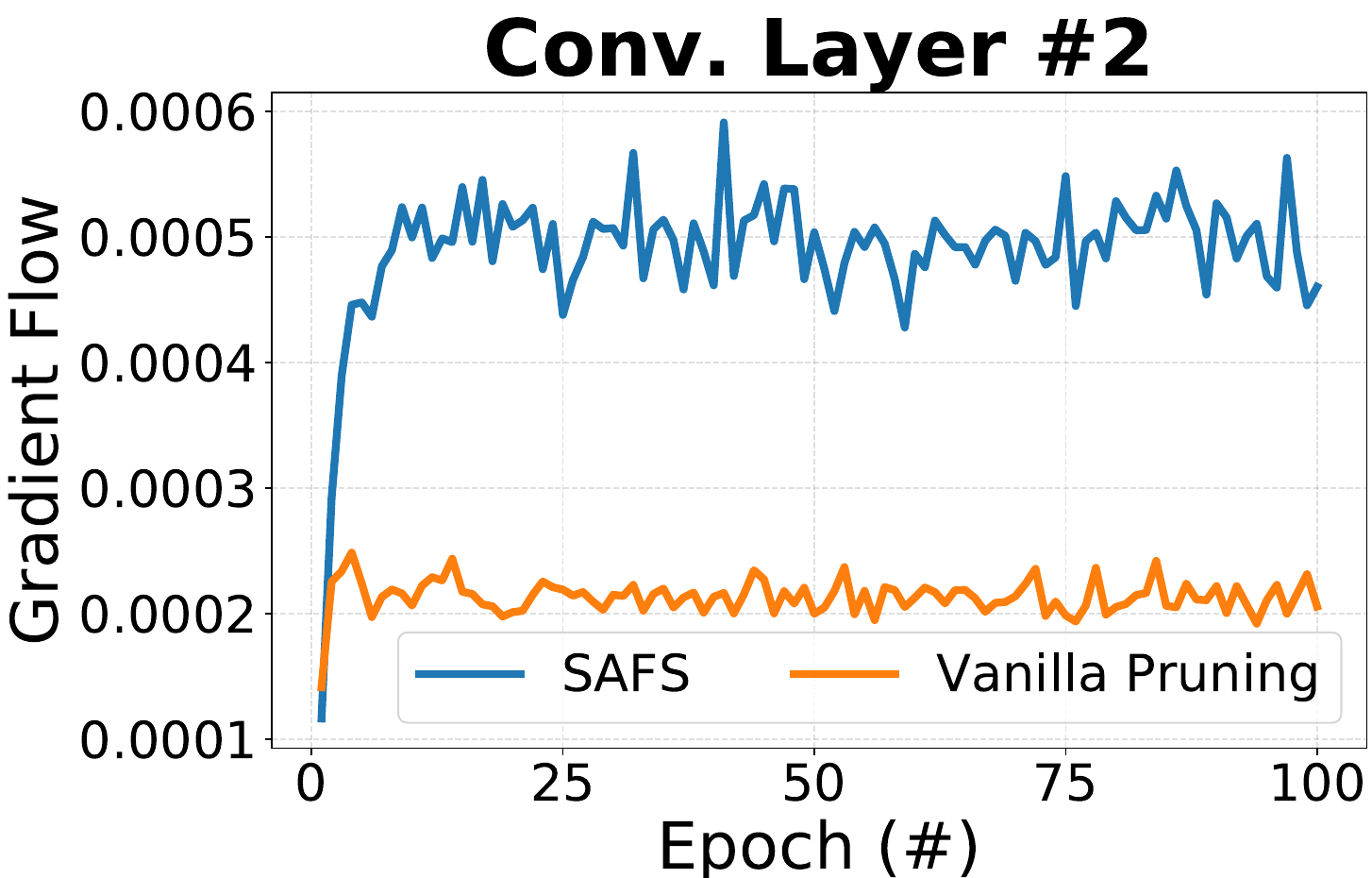}
\\
\includegraphics[width=0.2\textwidth]{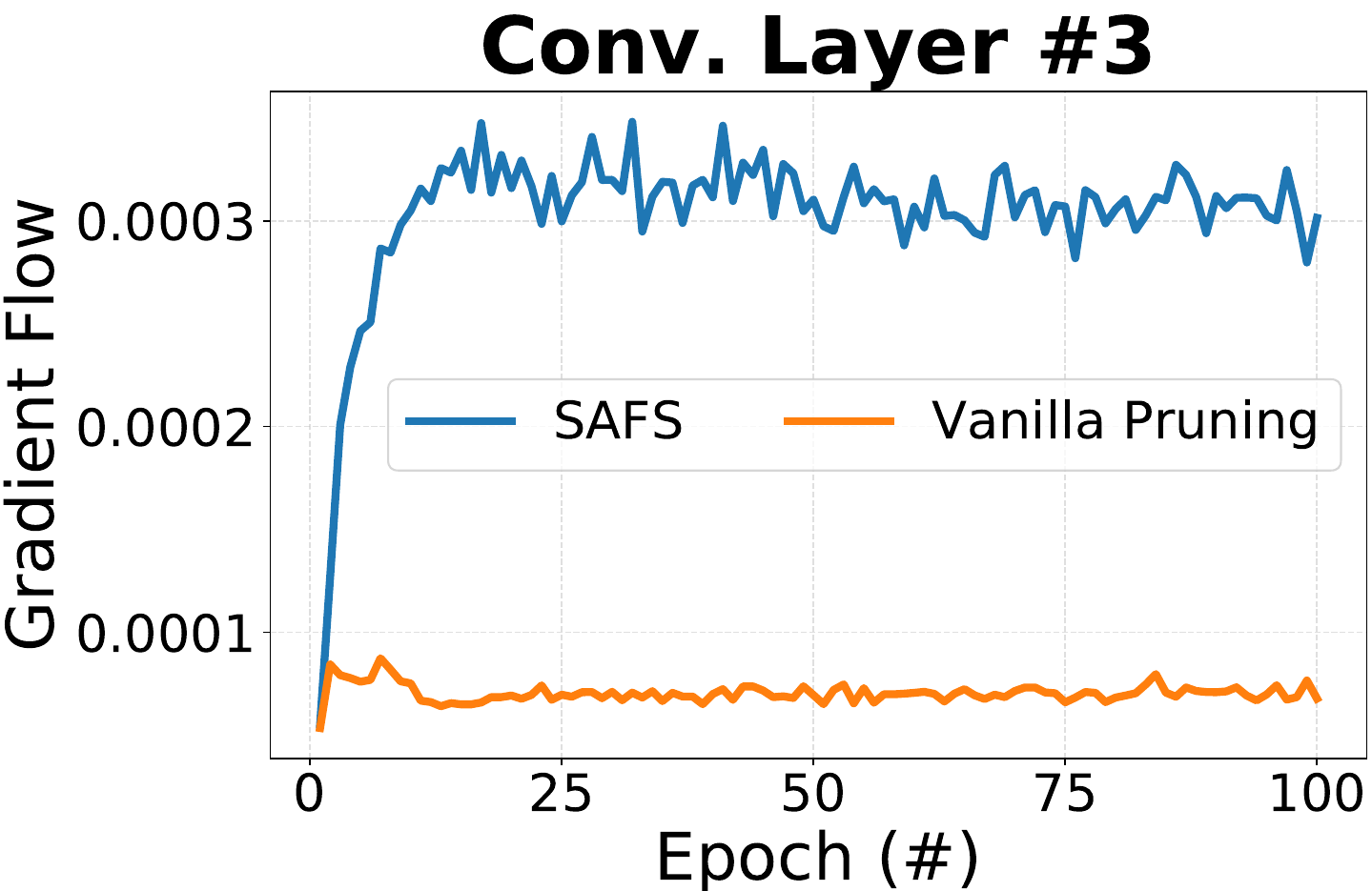}
&
\includegraphics[width=0.2\textwidth]{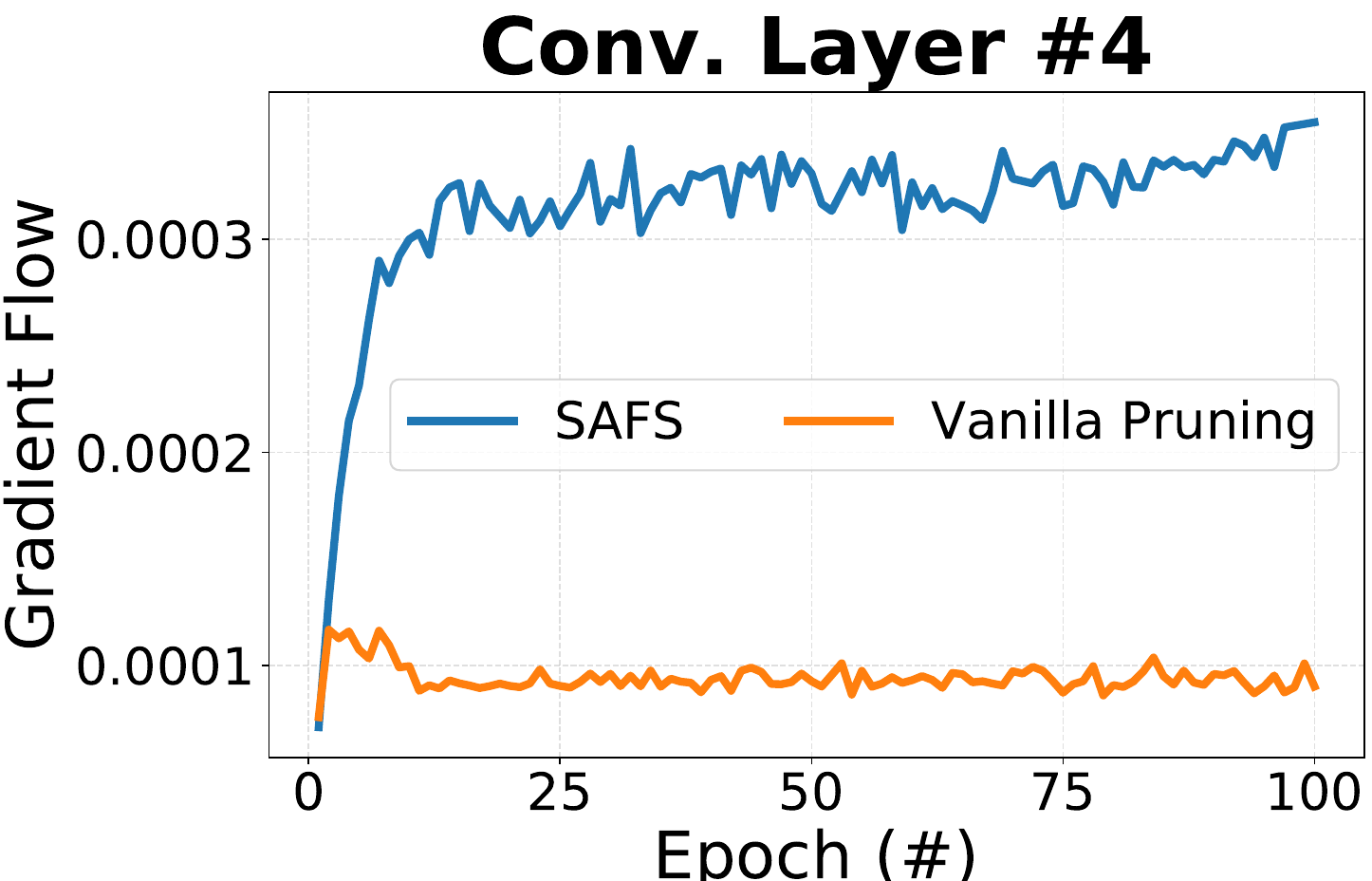}
\\
\multicolumn{2}{c}{\includegraphics[width=0.2\textwidth]{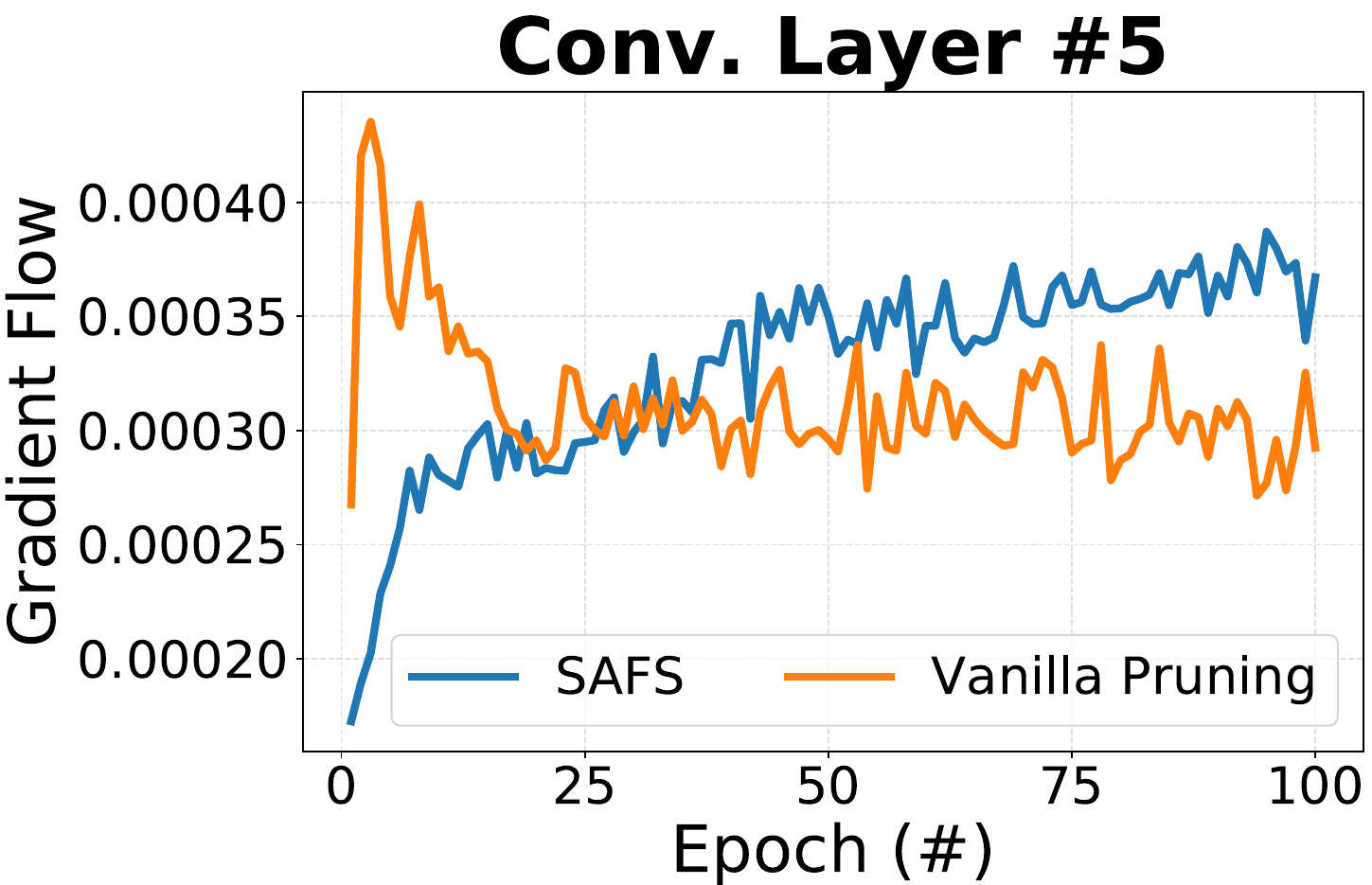}}
\end{tabular}
 }
\captionof{figure}{Gradient flow for sparse VGG-16 with 99\% pruning ratio with five convolutional layers trained on CIFAR-10. Note that the higher values are the better.} 
\label{fig:gradient-flow}
\end{table}

\section{Reporting the Computing Cost of \ourname}
\label{Appendix:Cost}

\cref{tab:experimental-setup:computing-cost} compares the computing cost (GPU hours) of refining a \snn with \ourname and default vanilla pruning. Although \ourname is slower than the vanilla pruning method, we need to pay this cost only once. Our results show that the significant improvements achieved by \ourname are worth paying this cost. It is important to note that we have not used any multi-fidelity techniques to speed up the first search stage, which is one reason for our slow speed. The use of search acceleration techniques will be explored in the future. 

\begin{table}[hbpt]
    \centering
    \caption{Reporting the required computing cost for learning \snn \afs.} 
    \label{tab:experimental-setup:computing-cost}
    \addtolength{\tabcolsep}{-3pt}
    \resizebox{0.85\textwidth}{!}{
    \begin{tabular}{c|c|c|c}
    \hline
    \multirow{3}{*}{\textbf{Network}} & \multirow{3}{*}{\textbf{Dataset}}  & \multicolumn{2}{c}{\textbf{GPU Hours (without considering dense training and sparsification)}}\\  \cline{3-4}
     &   &  \textbf{SAFS} & \textbf{Vanilla Pruning}\\
     &   &  \textbf{(with three-fold cross-validation)} & \textbf{(with one-fold cross-validation)}\\
    \hline
    LeNet-5 &  MNIST & 6.4 & 0.16\\
    VGG-16 &  CIFAR-10 & 47 & 3.8\\
    ResNet-18 & CIFAR-10  & 63 & 5.6 \\
    EfficientNet-B0 &  ImageNet-16 & 400 & 7.7 \\
    \hline
    \end{tabular}}
\end{table}

\section{Comparison of Accuracy-Compression Ratio Trade-off with State-of-the-Art}
\label{appendix:tradeoff}

\begin{figure}[htbp]
    \centering
    \begin{tikzpicture}[scale=1]
    \begin{axis}[
        ymin=0, ymax=10,
        xmin=0,xmax=110,
        grid=major, 
        grid style={dashed, gray!20},
        xlabel=\small	\textbf{Compression Ratio ($\times$) $\uparrow$ }, 
        ylabel=\small	 \textbf{Accuracy Improvement (\%) $\rightarrow$},
        ylabel near ticks,
        xlabel near ticks,
        legend style={at={(0.5,0.0)},anchor=north},
        legend style={draw=black, at={(1.51,1.)}, text opacity = 1,row sep=0.8pt, font=\fontsize{8}{6}\selectfont}, 
         x tick label style={rotate=0,anchor=north},
        ]
        
        \addplot [mark=oplus*, mark size=6pt, pink,  thick,draw=black] coordinates {(100,8.88) (20,7.45) (10,1.65)};

        \addplot [mark=diamond*, only marks, mark size=5pt, cyan, thick,draw=black] coordinates {(13.88,3)};

        \addplot [mark=square*, dashed, mark size=4pt, olive, thick, draw=black ] coordinates {(20,2.2)};

        \addplot [mark=square*, mark size=4pt, red, thick, draw=black] coordinates {(100,6.33)};

        \addplot [mark=x, mark size=5pt, only marks, gray, line width=2pt,draw=black] coordinates {(100,3.91)};

        \addplot [fill, mark=pentagon*, only marks, mark size=5pt, teal, thick,draw=black] coordinates {(7.14,1)};

        \addplot [fill, mark=halfcircle*, only marks, mark size=5pt, orange,draw=black] coordinates {(45.73,2.64)}; 

        \addplot [fill, mark= halfsquare left*, only marks, mark size=6pt, yellow,draw=black] coordinates {(16, 0.3)};

        \addplot [mark=triangle*, only marks, mark size=6pt, violet, thick,draw=black] coordinates {(20,4.4)};

        \addplot [dashed, ->, blue, line width=3pt] coordinates {(40,3.7)  (65,6.2)};
        \node[blue, rotate =43] at (axis cs: 48,5.6) {\large{\textbf{better result}}};
       
        \legend{VGG-16+\texttt{SAFS} (Ours), VGG-16+\citep{huang-wacv18}, VGG-16+Hydra \citep{sehwag-neurips20}, ResNet-18+\texttt{SAFS} (Ours), ResNet-18+Hydra \citep{sehwag-neurips20}, ResNet-18+One-Shot Pruning \citep{hubens-icip22}, ResNet-18+TAS \citep{loni-date22}, ResNet-18+TRQ \citep{li-aaai22}, DeepMaker \citep{loni-micro20}}
      \end{axis}
    \end{tikzpicture}
\caption{Showing the accuracy improvement (\%) vs. the number of network parameters (\#Params) of various compact networks trained on CIFAR-10.}
\label{fig:experiments:hardware}
\end{figure}
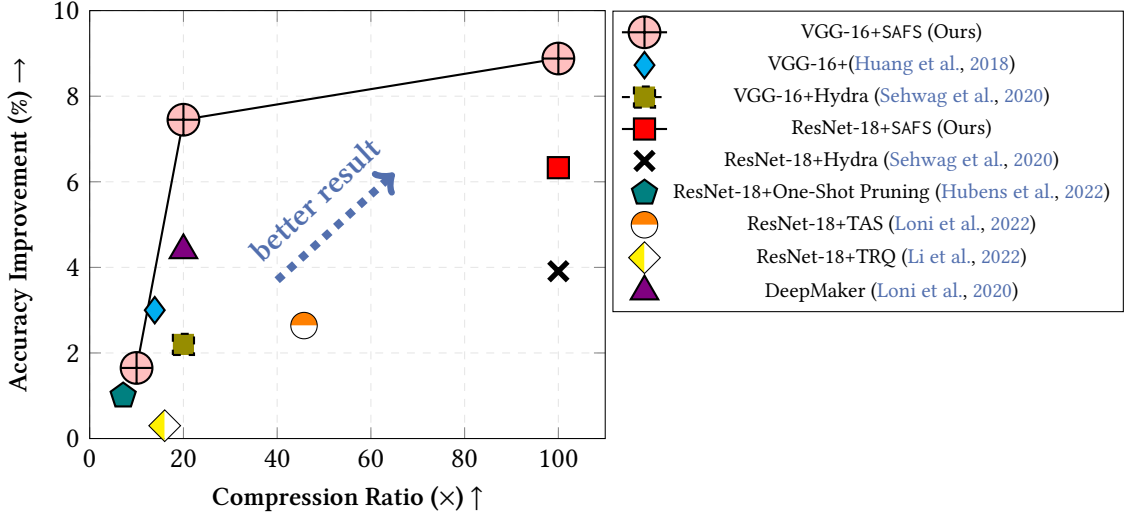

We study the effectiveness of \ourname in comparison with various state-of-the-art sparsification and quantization methods in the context of a trade-off between compression ratio ($x\mhyphen axis$) and performance improvement ($y\mhyphen axis$) compared to each method's baseline (\cref{fig:experiments:hardware}). We examine VGG-16 and ResNet-18 networks trained on CIFAR-10. 
Our results reveal that \ourname provides 6.24\% higher accuracy and 2.18$\times$ more compression ratio for VGG-16 over the best counterparts. \ourname achieves 2.42\% higher accuracy than the best counterparts with similar compression ratios for ResNet-18.

\section{Details on Searching Networks}
\label{appendix:search-details}
\cref{tab:search_details} shows the configuration details of Stage 1 and Stage 2 learning procedures. 

\begin{table}[htb]
\centering
\caption{Table showing the general hyperparameter configuration for \ourname learning procedures.}
\label{tab:search_details}
\resizebox{\textwidth}{!}{%
\begin{tabular}{cc}
     \hline
\multicolumn{2}{c}{\textbf{Stage 1: Learning Unary Operators} } \\ \hline
\multirow{3}{*}{Unary Operators$^\mp$}    
        & ReLU6 \citep{howard-corr17}, Acon \citep{ma-cvpr21},  TanhSoft-1 \citep{biswas-access21}\\ 
        &  SRS \citep{zhou-corr20}, Symlog \citep{hafner-corr23}, Symexp \citep{hafner-corr23}\\
        & Swish, Tanh, HardSwish, ELU, GELU, Softplus, LogisticSigmoid\\
    \hline
    History Length                      &  3\\
    Number of Iterations                & 20 \\ 
    Epochs for Evaluation               & 80\\ \hline
\multicolumn{2}{c}{\textbf{Stage 2: Scaling factors and HPO} } \\ \hline
    HPO Library & SMAC3$^\ast$\\
    Learning Rate & $1e^{-4}<$\textit{lr}$<1e^{-1}$ \\
    \multirow{2}{*}{Learning Rate Scheduler} & \textit{constant, step, linear, cosine annealing \citep{loshchilov-iclr17a}} \\ 
    & \textit{\{$0.001\times(0.5^{epoch\%20})$\}, ReduceLROnPlateau$^\dagger$, CosineAnnealingWarmRestarts$^\ddagger$}\\
    Optimizer &  SGD, Adam, Fromage, TAdam \citep{ilboudo-corr20} \\\hline
\multicolumn{2}{l}{$^\mp$ \citep{dubey-neurocomputing22} explains in detail popular \afs considered in this study. }\\
\multicolumn{2}{l}{$^\ast$ \url{https://github.com/automl/SMAC3}} \\ \multicolumn{2}{l}{$^\dagger$ \url{https://pytorch.org/docs/stable/generated/torch.optim.lr_scheduler.ReduceLROnPlateau.html}} \\
\multicolumn{2}{l}{$^\ddagger$ \url{https://pytorch.org/docs/stable/generated/torch.optim.lr_scheduler.CosineAnnealingWarmRestarts.html}} \\
\hline
\end{tabular}}
\end{table}

\cref{tab:experimental-setup:architecture-config:mnist} provides the configuration details for training the dense LeNet-5 model (baseline) with ReLU \afs trained on MNIST.

\begin{table}[hbpt]
    \centering
    \caption{Dense CNNs with training hyperparameters for MNIST dataset used in experiments.} 
    \label{tab:experimental-setup:architecture-config:mnist}
    \addtolength{\tabcolsep}{-3pt}
    \resizebox{\textwidth}{!}{
    \begin{tabular}{c|c}
    \hline
    \textbf{Network$^\ddagger$ }            & LeNet-5  \\ 
    \hline
    Epoch (\#)                  & 100      \\
    Learning Rate (\textit{lr}) & 0.1     \\
    Learning Rate Scheduler     &   None \\
    Optimizer                   &   SGD  \\ 
    \hline
    Train Time (GPU Hours) for One Model (One-fold) &  0.16\\ 
    \hline
     \multicolumn{2}{l}{$^\ddagger$ Original implementation of dense model: \url{https://github.com/ChawDoe/LeNet5-MNIST-PyTorch/blob/master/train.py}} \\ \hline
    
    \end{tabular}}
\end{table}

\cref{tab:experimental-setup:architecture-config:cifar} provides the configuration details for training dense models (baseline) with ReLU \afs trained on CIFAR-10. 

\begin{table}[hbpt]
    \centering
    \caption{Dense CNNs with training hyperparameters for CIFAR-10 dataset used in experiments.} 
    \label{tab:experimental-setup:architecture-config:cifar}
    \addtolength{\tabcolsep}{-3pt}
    \resizebox{\textwidth}{!}{
    \begin{tabular}{c|c|c}
    \hline
    \textbf{Network}            & VGG-16                                & ResNet-18 \\ 
    \hline
    Epoch (\#)                  & 200                                 & 200  \\
    Learning Rate (\textit{lr}) & 0.001                               &  0.01  \\
    \multirow{2}{*}{Learning Rate Scheduler}     & \multirow{2}{*}{\textit{$0.001\times(0.5^{epoch\%20}$)}}     &  \textit{ReduceLROnPlateau}$^\ddagger$: 
\\
& & \{factor: 0.05, patience: 2, min\_lr: 0, threshold: 0.0001, eps:1$e^{-8}$ \}
\\
    Weight Decay                & 5$e^{-4}$                                  &  5$e^{-4}$   \\ 
    Momentum                    & 0.9                                   & 0.9  \\ 
    Optimizer                   & SGD                                   & SGD  \\ 
    \hline
    Train Time (GPU Hours) for One Model (One-fold)  & 1.25                                   & 4.0 \\ 
    \hline
    \multicolumn{3}{l}{$^\ddagger$ \url{https://pytorch.org/docs/stable/generated/torch.optim.lr_scheduler.ReduceLROnPlateau.html}} \\ \hline
    \end{tabular}}
\end{table}

\cref{tab:experimental-setup:architecture-config:imagenet} provides the configuration details for training dense models (baseline) with ReLU \afs trained on ImageNet-16. 

\begin{table}[hbpt]
    \centering
    \caption{Dense CNNs with training hyperparameters for ImageNet-16 dataset used in experiments.} 
    \label{tab:experimental-setup:architecture-config:imagenet}
    \addtolength{\tabcolsep}{-3pt}
    \resizebox{\textwidth}{!}{
    \begin{tabular}{c|c|c}
    \hline
    \textbf{Network} & EfficientNet-B0  & ResNet-18 \\ 
    \hline
    Epoch (\#)                  &  50  & 50  \\
    Learning Rate (\textit{lr}) &  0.01 &  0.1  \\ \cline{2-3}
    \multirow{3}{*}{Learning Rate Scheduler}     &   \textit{CosineAnnealingWarmRestarts}$^\ddagger$: & \textit{CosineAnnealingWarmRestarts}$^\ddagger$: 
\\
& \{\#Iterations for first restart: 12, & \{\#Iterations for first restart: 12,
\\
&  Minimum learning rate:5$e^{-5}$ \}&  Minimum learning rate:5$e^{-5}$ \}
\\\cline{2-3}
    Weight Decay  & 5$e^{-4}$    &  5$e^{-4}$   \\ 
    Momentum   & 0.9 & 0.9  \\ 
    Optimizer   & SGD  & SGD  \\ 
    \hline
    Train Time (GPU Hours) for One Model (One-fold) &  18 & 16 \\ 
    \hline
    \multicolumn{3}{l}{$^\ddagger$ \url{https://pytorch.org/docs/stable/generated/torch.optim.lr_scheduler.CosineAnnealingWarmRestarts.html}} \\ \hline
    \end{tabular}}
\end{table}

\cref{tab:experimental-setup:architecture-config:hw-spec} presents specifications of hardware devices utilized for evaluating the performance of \ourname.

\begin{table}[hbpt]
    \centering
    \caption{Hardware Specification for search \& train.}
    \label{tab:experimental-setup:architecture-config:hw-spec}
    \addtolength{\tabcolsep}{-3pt}
    \begin{tabular}{c|c}
    \hline
    \textbf{Parameter}              & \textbf{Specification}\\ \hline
    GPU                             & NVIDIA\textsuperscript{®} RTX A4000 (735 MHz)\\
    GPU Memory                      & 16 GB GDDR6  \\
    GPU Compiler                    &  cuDNN version 11.1  \\ 
    System Memory                   & 64 GB \\
    Operating System                & Ubuntu 18.04 \\ 
    $CO_2$ Emission/Day $^\dagger$  & 1.45 Kg \\
    \hline 
    \multicolumn{2}{c}{$\dagger$ Calculated using the ML $CO_2$ impact framework \citep{lacoste-corr2019}.}	\\ 
    \hline
    \end{tabular}
\end{table}

%% file: main.bbl
\begin{thebibliography}{}

\bibitem[Apicella et~al., 2021]{apicella-nn20}
Apicella, A., Donnarumma, F., Isgr{\`o}, F., and Prevete, R. (2021).
\newblock A survey on modern trainable activation functions.
\newblock {\em Neural Networks}, 138.

\bibitem[Azarian et~al., 2020]{azarian-corr20}
Azarian, K., Bhalgat, Y., Lee, J., and Blankevoort, T. (2020).
\newblock Learned threshold pruning.
\newblock {\em CoRR}.

\bibitem[Basirat and Roth, 2021]{basirat-visigrapp21}
Basirat, M. and Roth, P. (2021).
\newblock S* relu: Learning piecewise linear activation functions via particle
  swarm optimization.
\newblock In {\em Proceedings of the 16th International Joint Conference on
  Computer Vision, Imaging and Computer Graphics Theory and Applications}.

\bibitem[Bingham et~al., 2020]{bingham-gecco20}
Bingham, G., Macke, W., and Miikkulainen, R. (2020).
\newblock Evolutionary optimization of deep learning activation functions.
\newblock In Ceberio, J., editor, {\em Proceedings of the Genetic and
  Evolutionary Computation Conference ({GECCO}'20)}. ACM Press.

\bibitem[Bingham and Miikkulainen, 2022]{bingham-NN22}
Bingham, G. and Miikkulainen, R. (2022).
\newblock Discovering parametric activation functions.
\newblock {\em Neural Networks}.

\bibitem[Biswas et~al., 2021]{biswas-access21}
Biswas, K., Kumar, S., Banerjee, S., and Pandey, A.~K. (2021).
\newblock Tanhsoft—dynamic trainable activation functions for faster learning
  and better performance.
\newblock {\em IEEE Access}, 9.

\bibitem[Burke and Bykov, 2017]{burke-ejor17}
Burke, E. and Bykov, Y. (2017).
\newblock The late acceptance hill-climbing heuristic.
\newblock {\em European Journal of Operational Research}.

\bibitem[Chaudhuri et~al., 2022]{icml22}
Chaudhuri, K., Jegelka, S., Song, L., Szepesvári, C., Niu, G., and Sabato, S.,
  editors (2022).
\newblock {\em Proceedings of the 39th International Conference on Machine
  Learning ({ICML}'22)}, volume 162 of {\em Proceedings of Machine Learning
  Research}. PMLR.

\bibitem[Chaudhuri and Salakhutdinov, 2019]{icml19}
Chaudhuri, K. and Salakhutdinov, R., editors (2019).
\newblock {\em Proceedings of the 36th International Conference on Machine
  Learning ({ICML}'19)}, volume~97. Proceedings of Machine Learning Research.

\bibitem[Chen and Ran, 2019]{chen-ieee2019}
Chen, J. and Ran, X. (2019).
\newblock Deep learning with edge computing: A review.
\newblock In {\em Proc. of the IEEE}.

\bibitem[Chrabaszcz et~al., 2017]{chrabaszcz-arxiv17a}
Chrabaszcz, P., Loshchilov, I., and Hutter, F. (2017).
\newblock A downsampled variant of {ImageNet} as an alternative to the {CIFAR}
  datasets.
\newblock {\em arXiv:1707.08819 {[cs.CV]}}.

\bibitem[Curci et~al., 2021]{curci-corr21}
Curci, S., Mocanu, D., and Pechenizkiyi, M. (2021).
\newblock Truly sparse neural networks at scale.
\newblock {\em CoRR}.

\bibitem[Dubey et~al., 2022]{dubey-neurocomputing22}
Dubey, S.~R., Singh, S.~K., and Chaudhuri, B.~B. (2022).
\newblock Activation functions in deep learning: A comprehensive survey and
  benchmark.
\newblock {\em Neurocomputing}, 503.

\bibitem[Dubowski, 2020]{dubowski-corr20}
Dubowski, A. (2020).
\newblock Activation function impact on sparse neural networks.
\newblock {\em CoRR}.

\bibitem[Evci et~al., 2020]{evci-icml20}
Evci, U., Gale, T., Menick, J., Castro, P., and Elsen, E. (2020).
\newblock Rigging the lottery: Making all tickets winners.
\newblock In \cite{icml20}.

\bibitem[Evci et~al., 2022]{evci-aaai22}
Evci, U., Ioannou, Y., Keskin, C., and Dauphin, Y. (2022).
\newblock Gradient flow in sparse neural networks and how lottery tickets win.
\newblock In \cite{aaai22}.

\bibitem[Evci et~al., 2019]{evci-corr19}
Evci, U., Pedregosa, F., Gomez, A., and Elsen, E. (2019).
\newblock The difficulty of training sparse neural networks.
\newblock {\em CoRR}.

\bibitem[Fedorov et~al., 2019]{fedorov-neurips19a}
Fedorov, I., Adams, R., Mattina, M., and Whatmough, P. (2019).
\newblock Sparse: Sparse architecture search for cnns on resource-constrained
  microcontrollers.
\newblock In \cite{neurips19}.

\bibitem[Frankle et~al., 2020]{frankle-icml20}
Frankle, J., Dziugaite, G., Roy, D., and Carbin, M. (2020).
\newblock Linear mode connectivity and the lottery ticket hypothesis.
\newblock In \cite{icml20}.

\bibitem[Godfrey, 2019]{godfrey-smc19}
Godfrey, L.~B. (2019).
\newblock An evaluation of parametric activation functions for deep learning.
\newblock In {\em {IEEE} International Conference on Systems, Man and
  Cybernetics}.

\bibitem[Graesser et~al., 2022]{graesser-icml22}
Graesser, L., Evci, U., Elsen, E., and Castro, P. (2022).
\newblock The state of sparse training in deep reinforcement learning.
\newblock In \cite{icml22}.

\bibitem[Hafner et~al., 2023]{hafner-corr23}
Hafner, D., Pasukonis, J., Ba, J., and Lillicrap, T. (2023).
\newblock Mastering diverse domains through world models.
\newblock {\em CoRR}.

\bibitem[Han et~al., 2015]{han-neurips15}
Han, S., Pool, J., Tran, J., and Dally, W. (2015).
\newblock Learning both weights and connections for efficient neural network.
\newblock In Cortes, C., Lawrence, N., Lee, D., Sugiyama, M., and Garnett, R.,
  editors, {\em Proceedings of the 28th International Conference on Advances in
  Neural Information Processing Systems ({N}eur{IPS}'15)}. Curran Associates.

\bibitem[Hayou et~al., 2019]{hayou-icml19}
Hayou, S., Doucet, A., and Rousseau, J. (2019).
\newblock On the impact of the activation function on deep neural networks
  training.
\newblock In \cite{icml19}.

\bibitem[Hinton et~al., 2015]{hinton-corr15}
Hinton, G., Vinyals, O., and Dean, J. (2015).
\newblock Distilling the knowledge in a neural network.
\newblock {\em CoRR}.

\bibitem[Hoefler et~al., 2021]{hoefler-jmlr21}
Hoefler, T., Alistarh, D., Ben-Nun, T., D, N., and Peste, A. (2021).
\newblock Sparsity in deep learning: Pruning and growth for efficient inference
  and training in neural networks.
\newblock {\em Journal of Machine Learning Research}.

\bibitem[Howard et~al., 2017]{howard-corr17}
Howard, A.~G., Zhu, M., Chen, B., Kalenichenko, D., Wang, W., Weyand, T.,
  Andreetto, M., and Adam, H. (2017).
\newblock Mobilenets: Efficient convolutional neural networks for mobile vision
  applications.
\newblock {\em CoRR}.

\bibitem[Huang et~al., 2018]{huang-wacv18}
Huang, Q., Zhou, K., You, S., and Neumann, U. (2018).
\newblock Learning to prune filters in convolutional neural networks.
\newblock In {\em 2018 IEEE Winter Conference on Applications of Computer
  Vision (WACV)}.

\bibitem[Hubens et~al., 2022]{hubens-icip22}
Hubens, N., Mancas, M., Gosselin, B., Preda, M., and Zaharia, T. (2022).
\newblock One-cycle pruning: Pruning convnets with tight training budget.
\newblock In {\em 2022 IEEE International Conference on Image Processing
  (ICIP)}.

\bibitem[Hutter et~al., 2019]{hutter-book19a}
Hutter, F., Kotthoff, L., and Vanschoren, J., editors (2019).
\newblock {\em Automated Machine Learning: Methods, Systems, Challenges}.
\newblock Springer.
\newblock Available for free at http://automl.org/book.

\bibitem[III and Singh, 2020]{icml20}
III, H.~D. and Singh, A., editors (2020).
\newblock {\em Proceedings of the 37th International Conference on Machine
  Learning ({ICML}'20)}, volume~98. Proceedings of Machine Learning Research.

\bibitem[Ilboudo et~al., 2020]{ilboudo-corr20}
Ilboudo, W., Eric, L., Kobayashi, T., and Sugimoto, K. (2020).
\newblock Tadam: A robust stochastic gradient optimizer.
\newblock {\em CoRR}.

\bibitem[Jaderberg et~al., 2014]{jaderberg-bmvc14}
Jaderberg, M., Vedaldi, A., and Zisserman, A. (2014).
\newblock Speeding up convolutional neural networks with low rank expansions.
\newblock In {\em British Machine Vision Conference, {BMVC}}.

\bibitem[Jaiswal et~al., 2022]{jaiswal-icml22}
Jaiswal, A., Ma, H., Chen, T., Ding, Y., and Wang, Z. (2022).
\newblock Training your sparse neural network better with any mask.
\newblock In \cite{icml22}.

\bibitem[Jin et~al., 2016]{jin-aaai16}
Jin, X., Xu, C., Feng, J., Wei, Y., Xiong, J., and Yan, S. (2016).
\newblock Deep learning with s-shaped rectified linear activation units.
\newblock In Schuurmans, D. and Wellman, M., editors, {\em Proceedings of the
  Thirtieth {AAAI} Conference on Artificial Intelligence ({AAAI}'16)}. {AAAI}
  Press.

\bibitem[Krizhevsky et~al., 2014]{krizhevsky-cifar}
Krizhevsky, A., Nair, V., and Hinton, G. (2014).
\newblock The cifar-10 dataset.
\newblock {\em online: http://www.cs.toronto.edu/kriz/cifar.html}, 55.

\bibitem[Kusupati et~al., 2020]{kusupati-icml20}
Kusupati, A., Ramanujan, V., Somani, R., Wortsman, M., Jain, P., Kakade, S.,
  and Farhadi, A. (2020).
\newblock Soft threshold weight reparameterization for learnable sparsity.
\newblock In \cite{icml20}.

\bibitem[Lacoste et~al., 2019]{lacoste-corr2019}
Lacoste, A., Luccioni, A., Schmidt, V., and Dandres, T. (2019).
\newblock Quantifying the carbon emissions of machine learning.
\newblock {\em CoRR}.

\bibitem[Larochelle et~al., 2020]{neurips20}
Larochelle, H., Ranzato, M., Hadsell, R., Balcan, M.-F., and Lin, H., editors
  (2020).
\newblock {\em Proceedings of the 33rd International Conference on Advances in
  Neural Information Processing Systems ({N}eur{IPS}'20)}. Curran Associates.

\bibitem[LeCun et~al., 1998]{lecun-ieee1998}
LeCun, Y., Bottou, L., Bengio, Y., and Haffner, P. (1998).
\newblock Gradient-based learning applied to document recognition.
\newblock In {\em Proc. of the IEEE}.

\bibitem[Lee et~al., 2020]{lee-iclr20}
Lee, N., Ajanthan, T., Gould, S., and Torr, P. (2020).
\newblock A signal propagation perspective for pruning neural networks at
  initialization.
\newblock In {\em Proceedings of the International Conference on Learning
  Representations ({ICLR}'20)}.
\newblock Published online: \url{iclr.cc}.

\bibitem[Lee et~al., 2018]{lee-corr18}
Lee, N., Ajanthan, T., Lee, J., and Torr, P. H.~S. (2018).
\newblock Snip: Single-shot network pruning based on connection sensitivity.
\newblock {\em CoRR}.

\bibitem[Li et~al., 2023]{li-corr23}
Li, G., Yang, Y., Bhardwaj, K., and Marculescu, R. (2023).
\newblock Zico: Zero-shot nas via inverse coefficient of variation on
  gradients.
\newblock {\em CoRR}.

\bibitem[Li et~al., 2016]{li-corr16}
Li, H., Kadav, A., Durdanovic, I., Samet, H., and Graf, H.~P. (2016).
\newblock Pruning filters for efficient convnets.
\newblock {\em CoRR}.

\bibitem[Li et~al., 2022]{li-aaai22}
Li, Y., Ding, W., Liu, C., Zhang, B., and Guo, G. (2022).
\newblock Trq: Ternary neural networks with residual quantization.
\newblock In \cite{aaai22}.

\bibitem[Lindauer et~al., 2022]{lindauer-jmlr22a}
Lindauer, M., Eggensperger, K., Feurer, M., Biedenkapp, A., Deng, D.,
  Benjamins, C., Ruhkopf, T., Sass, R., and Hutter, F. (2022).
\newblock {SMAC3}: A versatile bayesian optimization package for
  {H}yperparameter {O}ptimization.
\newblock {\em Journal of Machine Learning Research}, 23(54):1--9.

\bibitem[Liu et~al., 2018]{liu-corr2018}
Liu, Z., Sun, M., Zhou, T., Huang, G., and Darrell, T. (2018).
\newblock Rethinking the value of network pruning.
\newblock {\em CoRR}.

\bibitem[Loni et~al., 2022]{loni-date22}
Loni, M., Mousavi, H., Riazati, M., Daneshtalab, M., and Sj{\"o}din, M. (2022).
\newblock Tas: ternarized neural architecture search for resource-constrained
  edge devices.
\newblock In {\em 2022 Design, Automation \& Test in Europe Conference \&
  Exhibition (DATE)}.

\bibitem[Loni et~al., 2020]{loni-micro20}
Loni, M., Sinaei, S., Zoljodi, A., Daneshtalab, M., and Sj{\"o}din, M. (2020).
\newblock Deepmaker: A multi-objective optimization framework for deep neural
  networks in embedded systems.
\newblock {\em Microprocessors and Microsystems}, 73.

\bibitem[Loshchilov and Hutter, 2017]{loshchilov-iclr17a}
Loshchilov, I. and Hutter, F. (2017).
\newblock {SGDR}: Stochastic gradient descent with warm restarts.
\newblock In {\em Proceedings of the International Conference on Learning
  Representations ({ICLR}'17)}.
\newblock Published online: \url{iclr.cc}.

\bibitem[Ma et~al., 2021]{ma-cvpr21}
Ma, N., Zhang, X., Liu, M., and Sun, J. (2021).
\newblock Activate or not: Learning customized activation.
\newblock In {\em Proceedings of the International Conference on Computer
  Vision and Pattern Recognition ({CVPR}'21)}. Computer Vision Foundation and
  IEEE Computer Society, IEEE.

\bibitem[Mocanu et~al., 2018]{mocanu-nauture18}
Mocanu, D., Mocanu, E., Stone, P., Nguyen, P., Gibescu, M., and Liotta, A.
  (2018).
\newblock Scalable training of artificial neural networks with adaptive sparse
  connectivity inspired by network science.
\newblock {\em Nature Communications}.

\bibitem[Mostafa and Wang, 2019]{mostafa-icml19}
Mostafa, H. and Wang, X. (2019).
\newblock Parameter efficient training of deep convolutional neural networks by
  dynamic sparse reparameterization.
\newblock In \cite{icml19}.

\bibitem[Mousavi et~al., 2022]{mousavi-corr22}
Mousavi, H., Loni, M., Alibeigi, M., and Daneshtalab, M. (2022).
\newblock Pr-darts: Pruning-based differentiable architecture search.
\newblock {\em CoRR}.

\bibitem[Nair and Hinton, 2010]{nair-icml10}
Nair, V. and Hinton, G.~E. (2010).
\newblock Rectified linear units improve restricted boltzmann machines.
\newblock In F{\"u}rnkranz, J. and Joachims, T., editors, {\em Proceedings of
  the 27th International Conference on Machine Learning ({ICML}'10)}.
  Omnipress.

\bibitem[Nazari et~al., 2019]{nazari-dsd19}
Nazari, N., Loni, M., Salehi, M., Daneshtalab, M., and Sj\"odin, M. (2019).
\newblock Tot-net: An endeavor toward optimizing ternary neural networks.
\newblock In {\em 2019 22nd Euromicro Conference on Digital System Design
  (DSD)}.

\bibitem[Ramachandran et~al., 2018]{ramachandran-iclr18a}
Ramachandran, P., Zoph, B., and Le, Q. (2018).
\newblock Searching for activation functions.
\newblock In {\em Proceedings of the International Conference on Learning
  Representations ({ICLR}'18)}.
\newblock Published online: \url{iclr.cc}.

\bibitem[Sandler et~al., 2018]{sandler-cvpr18}
Sandler, M., Howard, A., Zhu, M., Zhmoginov, A., and Chen, L. (2018).
\newblock Mobilenetv2: Inverted residuals and linear bottlenecks.
\newblock In {\em Proceedings of the International Conference on Computer
  Vision and Pattern Recognition ({CVPR}'18)}. Computer Vision Foundation and
  IEEE Computer Society, IEEE.

\bibitem[Sehwag et~al., 2020]{sehwag-neurips20}
Sehwag, V., Wang, S., Mittal, P., and Jana, S. (2020).
\newblock Hydra: Pruning adversarially robust neural networks.
\newblock In \cite{neurips20}.

\bibitem[Selman and Gomes, 2006]{selman-encyclopedia06}
Selman, B. and Gomes, C. (2006).
\newblock Hill-climbing search.
\newblock In {\em Encyclopedia of cognitive science}.

\bibitem[Simonyan and Zisserman, 2014]{simonyan-arxiv14a}
Simonyan, K. and Zisserman, A. (2014).
\newblock Very deep convolutional networks for large-scale image recognition.
\newblock {\em arXiv:1409.1556 [cs.CV]}.

\bibitem[Srinivas et~al., 2017]{srinivas-ieee17}
Srinivas, S., Subramanya, A., and Babu, R. (2017).
\newblock Training sparse neural networks.
\newblock In {\em IEEE Conference on Computer Vision and Pattern Recognition
  Workshops, {CVPR} Workshops 2017, Honolulu, HI, USA, July 21-26}.

\bibitem[Subramanian et~al., 2022]{subramanian-bigdata22}
Subramanian, M., Lv, N.~P., and VE, S. (2022).
\newblock Hyperparameter optimization for transfer learning of vgg16 for
  disease identification in corn leaves using bayesian optimization.
\newblock {\em Big Data}, 10.

\bibitem[Sycara et~al., 2022]{aaai22}
Sycara, K., Honavar, V., and Spaan, M., editors (2022).
\newblock {\em Proceedings of the Thirty-Sixth Conference on Artificial
  Intelligence ({AAAI}'22)}. Association for the Advancement of Artificial
  Intelligence, {AAAI} Press.

\bibitem[Tanaka et~al., 2020]{tanaka-neurips20}
Tanaka, H., Kunin, D., Yamins, D.~L., and Ganguli, S. (2020).
\newblock Pruning neural networks without any data by iteratively conserving
  synaptic flow.
\newblock In \cite{neurips20}.

\bibitem[Tavakoli et~al., 2021]{tavakoli-nn21}
Tavakoli, M., Agostinelli, F., and Baldi, P. (2021).
\newblock Splash: Learnable activation functions for improving accuracy and
  adversarial robustness.
\newblock {\em Neural Networks}.

\bibitem[Tessera et~al., 2021]{tessera-corr21}
Tessera, K., Hooker, S., and Rosman, B. (2021).
\newblock Keep the gradients flowing: Using gradient flow to study sparse
  network optimization.
\newblock {\em CoRR}.

\bibitem[Vischer et~al., 2022]{vischer-iclr22}
Vischer, M., Lange, R., and Sprekeler, H. (2022).
\newblock On lottery tickets and minimal task representations in deep
  reinforcement learning.
\newblock In {\em Proceedings of the International Conference on Learning
  Representations ({ICLR}'22)}.
\newblock Published online: \url{iclr.cc}.

\bibitem[Wallach et~al., 2019]{neurips19}
Wallach, H., Larochelle, H., Beygelzimer, A., d'Alche Buc, F., Fox, E., and
  Garnett, R., editors (2019).
\newblock {\em Proceedings of the 32nd International Conference on Advances in
  Neural Information Processing Systems ({N}eur{IPS}'19)}. Curran Associates.

\bibitem[Zamora et~al., 2022]{zamora-aaai22}
Zamora, J., Rhodes, A.~D., and Nachman, L. (2022).
\newblock Fractional adaptive linear units.
\newblock In \cite{aaai22}.

\bibitem[Zhan and Cao, 2019]{zhan-corr19}
Zhan, H. and Cao, Y. (2019).
\newblock Deep model compression via deep reinforcement learning.
\newblock {\em CoRR}.

\bibitem[Zhou et~al., 2017]{zhou-iclr17}
Zhou, A., Yao, A., Guo, Y., Xu, L., and Chen, Y. (2017).
\newblock Incremental network quantization: Towards lossless cnns with
  low-precision weights.
\newblock In {\em Proceedings of the 5th International Conference on Learning
  Representations ({ICLR}'17)}.

\bibitem[Zhou et~al., 2019]{zhou-neurips19}
Zhou, H., Lan, J., Liu, R., and Yosinski, J. (2019).
\newblock Deconstructing lottery tickets: Zeros, signs, and the supermask.
\newblock In \cite{neurips19}.

\bibitem[Zhou et~al., 2020]{zhou-corr20}
Zhou, Y., Li, D., Huo, S., and Kung, S. (2020).
\newblock Soft-root-sign activation function.
\newblock {\em CoRR}.

\end{thebibliography}
